\documentclass[twoside,leqno,twocolumn, 10pt]{article}

\usepackage[letterpaper]{geometry}
\usepackage[subtle]{savetrees}
\usepackage{ltexpprt}
\usepackage{hyperref}
\usepackage{lipsum}
\usepackage{amsmath,amsfonts}
\usepackage{algorithmic} 
\usepackage{textcomp}
\usepackage{xcolor}
\usepackage{subfiles}
\usepackage{caption}
\usepackage{subcaption}
\usepackage[symbol]{footmisc}
\usepackage[algo2e, ruled, vlined, linesnumbered]{algorithm2e}

\usepackage{array}
\usepackage{multirow}
\usepackage{titlesec}
\usepackage{enumitem} 
\usepackage{tikz}
\usepackage{pifont}
\usepackage{array}
\usepackage{fontawesome}
\usepackage{framed}

\newtheorem{definition}{Definition}
\newcommand{\cmark}{\ding{51}}%

\definecolor{blue}{rgb}{.25,.25,.70}

\newcolumntype{P}[1]{>{\centering\arraybackslash}p{#1}}

\def\BibTeX{{\rm B\kern-.05em{\sc i\kern-.025em b}\kern-.08em
    T\kern-.1667em\lower.7ex\hbox{E}\kern-.125emX}}

\begin{document}
\title{G2A2: An Automated Graph Generator \\ with Attributes and Anomalies}

\author{Saikat Dey\thanks{Virginia Tech, Blacksburg, VA, USA  Email: \{dsaikat, sonalj, wfeng\}@vt.edu}
\and Sonal Jha\protect\protect\footnotemark[1]
\and Wu-chun Feng\protect\protect\footnotemark[1]
}
\date{}

\maketitle

\begin{abstract}
Many data-mining applications use dynamic attributed graphs to
represent relational information; but due to security and privacy concerns, there is a dearth of 
available datasets that can be represented as dynamic attributed graphs.
Even when such datasets are available, they do \emph{not} have ground truth that can be used to train deep-learning models. Thus, we present \textsf{G2A2}, an automated \underline{g}raph \underline{g}enerator with \underline{a}ttributes and \underline{a}nomalies, which encompasses (1) probabilistic models to generate a dynamic bipartite graph, representing time-evolving connections between two independent sets of entities, (2) realistic injection of anomalies using a novel algorithm that captures the general properties of graph anomalies across domains, and (3) a deep generative model to produce realistic attributes, learned
from an existing real-world dataset. 
Using the maximum mean discrepancy (MMD) metric to evaluate the realism of a
\textsf{G2A2}-generated graph against three real-world graphs,
\textsf{G2A2} outperforms Kronecker graph generation by reducing the MMD distance by up to six-fold (6$\times$). 
\end{abstract}


\section{Introduction} \label{Introduction}

Dynamic attributed graphs represent relational information in many data-mining applications (across various domains), including fraud detection (commerce), intrusion detection (networking), and recommendation systems (social media). However, because of the
security, privacy, and obfuscation of data, there is a paucity of available datasets that can be represented as dynamic attributed graphs.
Even when such datasets are available, there is \emph{no ground truth} that can be used to train supervised deep-learning (DL) models. Furthermore, the size and structure of the associated graphs are insufficient to properly train DL models.
Even rarer are publicly available \emph{time-series} datasets with additional descriptive \emph{attributes} and \emph{ground truth} to enable the construction of dynamic attributed graphs that are useful for classification problems, e.g., anomaly detection~\cite{anomaly1}.
Consequently, DL researchers generate synthetic graphs to fill this void, but the graphs are \emph{not} realistic and often differ enough from real-world graphs that they result in poorly-trained DL models.
For example, while statistical graph generation models, e.g., Erdos-Renyl (E-R)~\cite{ermodel} and Kronecker~\cite{kroneckergraph}, are widely used to generate static and time-evolving graphs, they are overly simplistic and fail to capture the intricate patterns of a realistic time-evolving graph, e.g., the node and edge distribution over time. Moreover, they  neither generate realistic \emph{attributes} nor a \emph{ground truth} for classification problems. 
Subsequently, while robust statistical models, e.g., multiplicative attribute graph (MAG)~\cite{mag} model and DL methodologies like D2G2~\cite{d2g2}, can generate realistic graphs with attributes, the models require a significant number of dynamic attributed graphs to train on. Furthermore, such models cannot generate a ground truth or special graphs, e.g., bipartite graphs that represent the connections between two independent sets of entities (e.g., users and items), 
commonly found in domains like social media. To address the above shortcomings, we propose a methodology to generate realistic, dynamic, attributed, bipartite graphs with known instances of anomalies (i.e., ground truth), as encompassed by \textbf{G2A2}, \textit{our \underline{g}raph \underline{g}enerator with \underline{a}ttributes and \underline{a}nomalies}, which uses a hybrid statistical+DL approach. Table~\ref{tab:methodComparison} summarizes how our approach compares to the existing state of the art.
\begin{table}[tbh]
\centering
\caption{Comparison of graph-generation methods}
\scalebox{0.60}{
\renewcommand{\arraystretch}{1}
\begin{tabular}{|P{0.11\textwidth}|P{0.08\textwidth}|P{0.09\textwidth}|P{0.11\textwidth}|P{0.09\textwidth}|P{0.1\textwidth}|P{0.07\textwidth}|}
\hline
\textbf{Methods}                    & \textbf{Realism}      & \textbf{Dynamic}          & \textbf{Attributed}       & \textbf{Bipartite}    & \textbf{Without training}     & \textbf{Ground truth}            \\ \hline \hline
E-R~\cite{ermodel}                  &                       &                           &                           & \cmark                &   \cmark                      &                                            \\ \hline
Kronecker~\cite{kroneckergraph}     & \cmark                &  \cmark                   &                           & \cmark                &   \cmark                      &                                            \\ \hline
D2G2~\cite{d2g2}                    & \cmark                &    \cmark                       &     \cmark                &                       &                               &                                            \\ \hline
MAG~\cite{mag}                      & \cmark                & \cmark                    &     \cmark                &  \cmark               &   \cmark                      &                                            \\ \hline
\textbf{G2A2}                       & \textbf{\cmark}       & \textbf{\cmark}           &     \cmark                & \textbf{\cmark}       & \textbf{\cmark}               &    \textbf{\cmark}                        \\ \hline
\end{tabular}
}
\label{tab:methodComparison}
\end{table}

To the best of our knowledge, our methodology is the first to generate a \emph{synthetic graph that is realistic, dynamic, attributed, and bipartite with known instances of anomaly (i.e., ground truth)}. In all, our contributions are as follows:
\vspace*{-3pt}
\begin{itemize}[leftmargin=*]
    \itemsep0em
    \item G2A2, a \underline{g}raph \underline{g}enerator with \underline{a}ttributes and \underline{a}nomalies that produces realistic, dynamic, attributed, bipartite graphs with known instances of anomalies. 
    \begin{itemize}
    \vspace*{-2pt}
    \item A novel approach to inject anomalies with the properties of anomaly propagation and burstiness.
    \end{itemize}
    \item A rigorous evaluation of the quality and computation time of our generated synthetic graph compared to three real-world graph datasets (Reddit~\cite{jodie},  Wikipedia~\cite{jodie}, and P-core network traffic~\cite{pcore}) with respect to three similarity criteria: graph similarity, anomaly similarity, and attribute similarity.
    \item A realistic graph library containing social media graphs, article graphs, and internet traffic graphs with different anomaly percentages and number of nodes and edges, as generated by G2A2. The library can be accessed from here: \href{https://github.com/Sonalj96/G2A2-Graph-Library}{\textbf{https://github.com/Sonalj96/G2A2-Graph-Library}}
\end{itemize}
The rest of the paper is organized as follows: \S2 Methodology, \S3 Results and Evaluation, \S4 Related Work, and \S5 Conclusion and Future Work.
\section{Methodology} \label{Method} 

\begin{figure}[tbh]
            \centering
            \includegraphics[scale=0.52]{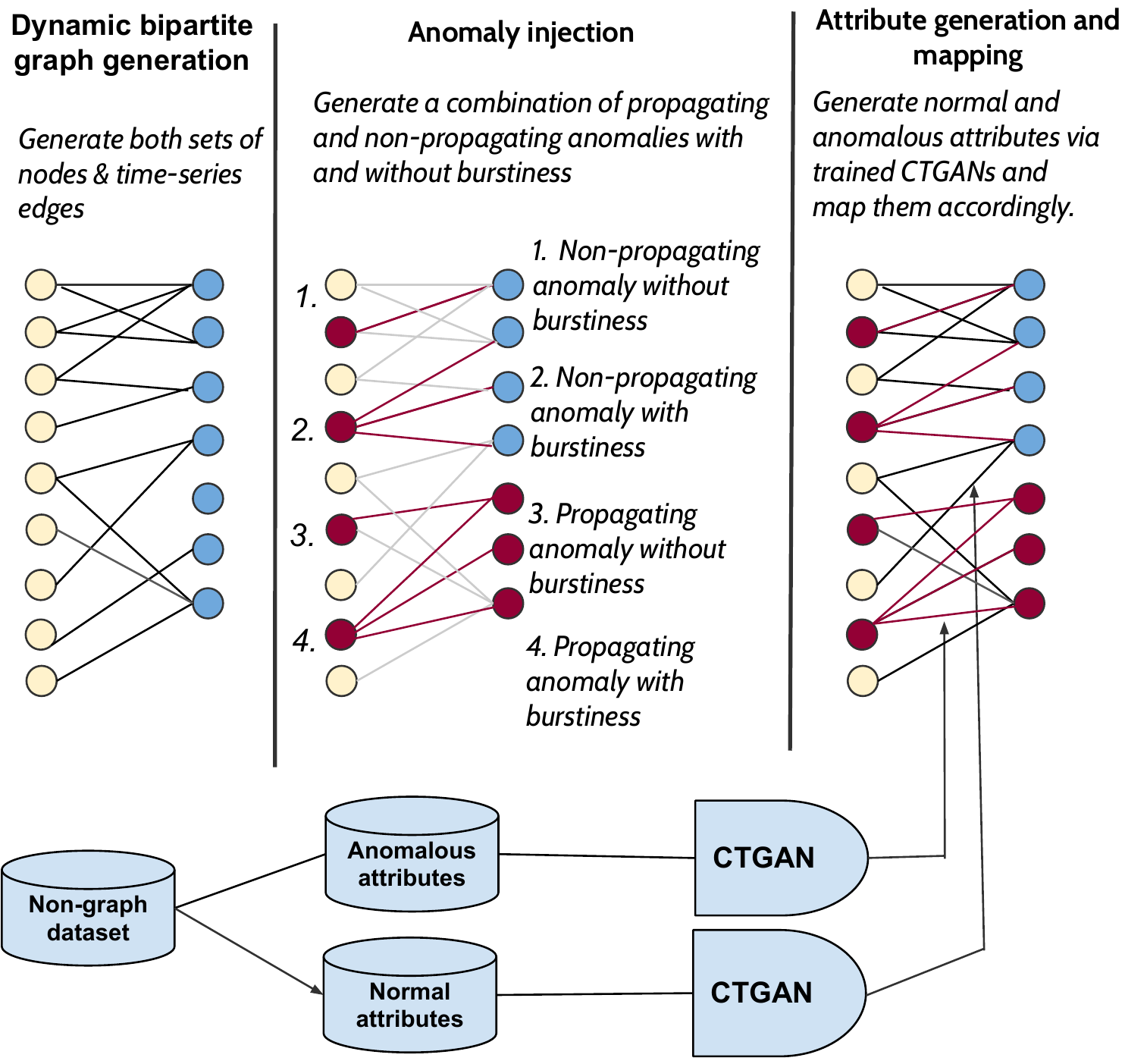}
            \caption{G2A2 methodology}
             \label{fig:workflow}
            \end{figure}

Figure~\ref{fig:workflow} illustrates our G2A2 methodology in three steps: 
(1) dynamic bipartite graph generation, 
(2) anomaly injection, and 
(3) attribute generation and mapping.
Our dynamic bipartite graph generation uniquely models time and degree distributions to generate realistic graphs. This is in contrast to the existing graph generation methodologies, which suffer from 
(1) modeling the entire graph under one distribution \emph{without} considering the time distribution of the graph snapshots and (2) adding only noise (or outliers) to the graph rather than adding anomalies.
Next, our novel anomaly injection algorithm emulates common graph-anomaly properties, such as burstiness~\cite{burstiness} and propagation~\cite{netwalk}, in the generated graph. Finally, by leveraging deep generative models like CTGAN~\cite{ctgan}, we generate and map realistic attributes to the graph. 

To explain our G2A2 methodology, we first define our notation for dynamic attributed bipartite graphs and anomalies as well as formulate our problem statement.

\begin{definition}
\textbf{Dynamic Attributed Bipartite Graph}. An attributed bipartite graph 
is a graph $G(U, V, E, F)$, where $U$ and $V$ are two non-overlapping sets of nodes, $E$ represents the edges connecting $U$ and $V$ (and there exists no edge within the sets themselves), and $F$ denotes the edge attributes or features of the graph $G$.
A dynamic or time-evolving attributed bipartite graph 
is a set of snapshots $G = \{G_1, G_2, G_3, .... G_t\}$ where $t \in T$, the total number of snapshots. 
\end{definition}
\begin{definition}
\vspace*{-3pt}
\textbf{Graph Anomalies}. Anomalies are rare occurrences or events in data that are often hard to detect. Given a graph $G$, anomalies constitute nodes or edges that are significantly different from the rest of the graph. In a graph $G$, we represent anomalies as $A = \{A_1, A_2, ... A_{r}\}$ where $r$ is the total number of injected anomalies. $A^U$ and $A^V$ are the sets of node anomalies, and $A^e$ represents the edge anomalies.
\end{definition}
\vspace*{-2pt}
Unlike an outlier that is just noise, anomalies have patterns associated with real-world events~\cite{anomalySurvey2}. 
For example, a distributed denial-of-service (DDoS) attack~\cite{ddos} is an anomaly.
In the context of a graph, there exist three types of anomalies; point-based, time-based~\cite{anomalySurvey2}, and graph-based~\cite{graphAnomaly}. Detecting
point-based anomalies can occur independently without exploring their relationship with the other data points. Time-based anomalies exhibit patterns over time; the best way to detect them is by plotting and viewing
individual anomalies over time. Graph-based anomalies, however, require a deeper understanding of the entities involved and the relationships between them.  They can only be detected when we view the relationships within a graph. 
\begin{definition}
\textbf{Problem Statement}.  
Given a non-graph dataset D, how to generate a dynamic attributed bipartite graph $G(U, V, E, F)$ with T snapshots and a set of graph anomalies $A = \{A_1, A_2,... A_{r}\}$? Graph G should have realistic statistical distributions (time and degree), and the attributes $F$ should be similar to D.   
\end{definition} 
\subsection{\textbf{Dynamic bipartite graph generation:}}
To generate a realistic dynamic bipartite graph, we need to model both a realistic time distribution and degree distribution. 
For the former, past studies show that the Gaussian and Cauchy distributions provide a realistic representation of the time-series data~\cite{cauchybetter}. For the latter, the degree distribution is often assumed to follow the power law~\cite{powerlaw}. However, recent studies show that the power-law distribution itself might be too narrow for modeling a realistic degree distribution; instead, a more general distribution, such as gamma, is more suitable~\cite{gammabetter}. 
Our experiments show that the Cauchy distribution best fits a realistic time distribution and that the gamma distribution works best for a realistic degree distribution. Below we explain 
how G2A2 generates a dynamic bipartite graph using the Cauchy and gamma distributions.  

\vspace*{9pt}
\noindent\textbf{Time Distribution.} 
            The frequency of specific events changes with time and often follows a cyclic pattern, i.e., repeats itself after $x$ units of time. For example, Internet traffic increases during the day and falls overnight
            every 24 hours. Based on our 
            experiments on diverse datasets across many domains, we found that the Cauchy distribution fits the best across such cyclic time patterns~\cite{cauchy2}.
            In addition, we found that the number of nodes participating in a graph snapshot also varies cyclically with time. 
            The Cauchy distribution is heavy-tailed~\cite{nist}, as shown in Equation~(\ref{eq:cauchy}):
            \begin{equation}
            \label{eq:cauchy}
            f(x) = \dfrac{1}{(s\pi(1+(x-l)s)^2)}
            \end{equation}
            where $l$ is the location parameter and $s$ is the scale parameter.
            
            The general idea is to use the Cauchy distribution to generate the time distribution for $x$ hours and then repeat
            it $T/x$ times, where $T$ is the total number of graph snapshots. By combining all the $T/x$ Cauchy distributions, we obtain the overall time distribution. We calculate three such time distributions for edges $E$, nodes $U$, and nodes $V$ and then use these 
            time 
            distributions to sample nodes and calculate the number of edges for a single graph snapshot. Using the obtained edge count, we obtain the degree sequence of the sampled nodes via the gamma distribution.
            
            To sample the nodes, we use weighted sampling~\cite{weightsampling}, where the weight is inversely proportional to the node degree probability obtained from the gamma distribution. 
            We then feed the degree sequence as input to the configuration model to get the final graph snapshot. This process iterates to generate all the graph snapshots in the graph, as articulated in Algorithm~\ref{alg:gg}. 

            \setlength{\textfloatsep}{0pt}

\begin{algorithm2e}[tbh]
\SetAlgoLined
\KwIn{Cauchy (cPara) and gamma (gPara) parameters; total time ($T$); total number of U and V nodes ($|U|$, $|V|$); total number of edges ($|E|$) } 
\KwOut{Dynamic Bipartite Graph (G)}
\SetKwInOut{KwIn}{Input}
\SetKwInOut{KnOut}{Output}
$ U^{count} = Cauchy(|U|, T, cPara)$\; 
$ V^{count} = Cauchy(|V|, T, cPara)$\; 
$ E^{count} = Cauchy(|E|, T, cPara)$\; 
$ U^{probability} = Gamma(|U|, gPara)$ \; 
$ V^{probability} = Gamma(|V|, gPara)$ \; 

   \For{t in T}{ 
     $U^{snapshot} = sampleNodes(U^{count}_{t}, U^{probability})$\;
     $V^{snapshot} = sampleNodes(V^{count}_{t}, V^{probability})$\;
     $seq^{snapshot} = sampleEdges(U^{snapshot}, V^{snapshot}, E^{count}_{t})$\;
     $add(G_{t}, BiCM(seq^{snapshot}))$\;
  }
\caption{Dynamic Bipartite Graph Generation}
\label{alg:gg}
\end{algorithm2e}

            \vspace*{9pt}
            \noindent\textbf{Degree Distribution}:
            The degree distribution of any graph defines its primary structure. 
            Similar to our experiments performed on real-world datasets to deduce time distributions, we found that most real-world graph degrees have a long-tailed distribution~\cite{kroneckergraph}. 
            Out of the many different types of long-tailed distributions, 
            the gamma distribution fits the best across real-world datasets. The gamma distribution is a two-parameter family of continuous probability distributions~\cite{nist} represented by the following equation:
            \vspace*{-4pt}
            \begin{equation}
            g(x) = \dfrac{(\dfrac{(x-l)}{s})^{a - 1}\exp{(\dfrac{- (x - l)}{s})}}{s \Gamma(a)}   
            \end{equation}
            where $a$ is the shape parameter, $l$ is the location parameter, $s$ is the scale parameter, and $\Gamma$ is the gamma function, i.e.,
            \begin{equation}
            \Gamma \left ( a \right ) = \int_{ 0}^{\infty} t^{a-1}e^{-t}dt
            \end{equation}
            We calculate the degree distribution of both $U$ and $V$ nodes separately as they are disjoint sets. Next, we use this probability degree distribution to calculate the degree sequences of both the $U$ and $V$ nodes of the graph snapshot. Then we input these sequences to the bipartite configuration model (BiCM)~\cite{bicm}, as shown in  Algorithm~\ref{alg:gg}.
            
\begin{figure}[tbh]
                \centering
                \includegraphics[scale=0.4]{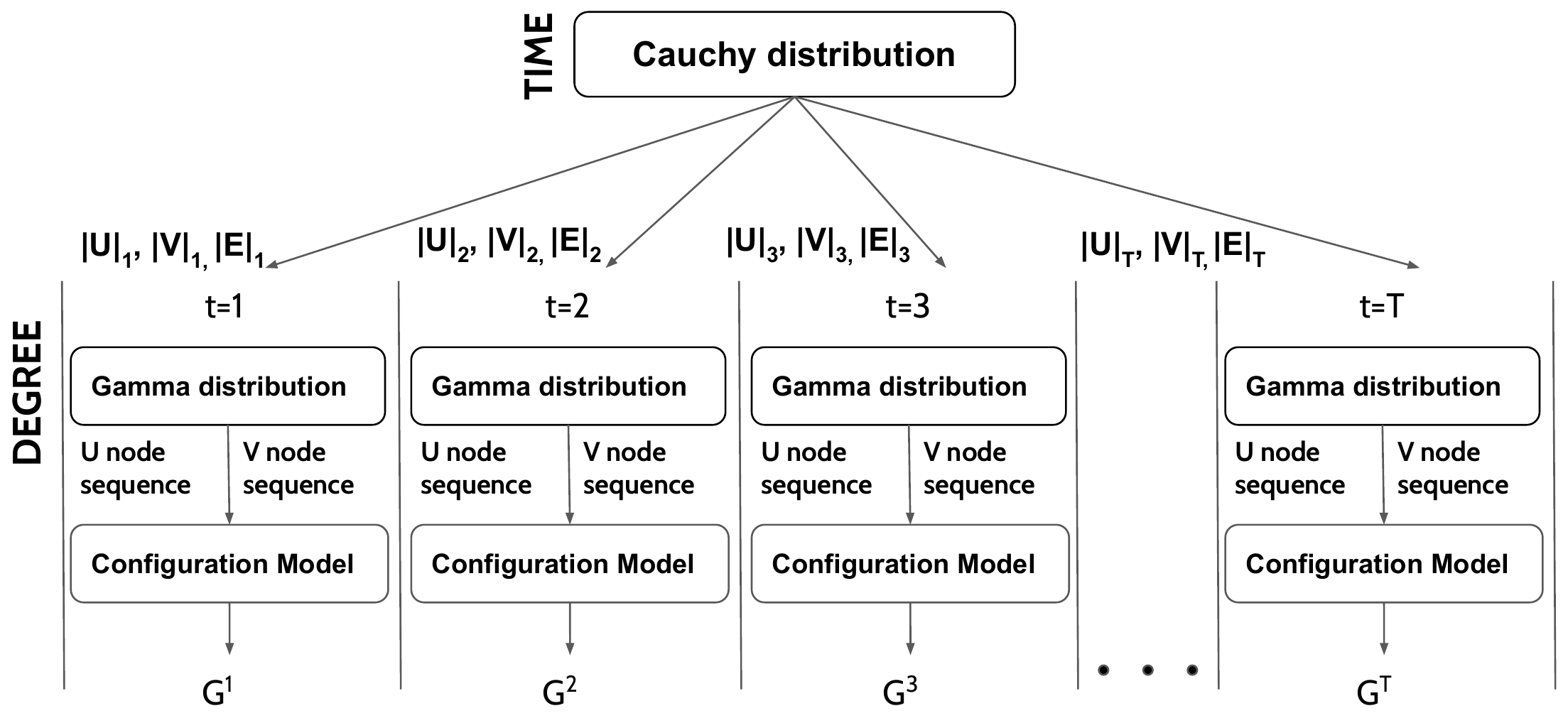} 
                \caption{Dynamic bipartite graph generation}
                \label{fig:DBGG}
            \end{figure}

            \vspace*{9pt}
            \noindent\textbf{Bipartite Configuration Model (BiCM)}:
            The BiCM generates a graph from a given degree sequence. The generated graph also has real-world graph properties, such as the small-world effect~\cite{smallworld} and high clustering. We use the bipartite version of the configuration model (BiCM)~\cite{bicm}, where we pass two degree sequences of $U$ and $V$ nodes, respectively, as inputs to the model. 
            Figure~\ref{fig:DBGG} shows a visualization of our dynamic bipartite graph generation.
            
\subsection{\textbf{Anomaly injection:}}
Here we seek to inject a range of graph anomalies into our graph based on some of the most common properties of graph anomalies across domains, such as burstiness and propagation. In a graph, anomalies can occur at a node or an edge;
however, edge anomalies are directly dependent on the change in the state of the nodes. For example, in network intrusion, when a system (represented by a node) gets compromised, unknown or suspicious transactions are made from that system (represented by anomalous edges) that the system user would not make otherwise. Additionally, the change in the state of the node can happen either due to a graph-associated event that occurred at a previous time step or some other external factors. For instance, a system can get compromised by an attack that was propagated from a previous connection, or the system can be the original attacker itself~\cite{mirai1}. 
The anomalous behavior differs between domains, too. For example, in 
social media fraud, anomalies do \emph{not} propagate, and the frequency of their occurrence is less than that of a malicious system attacking other systems in a network graph to gain access.

\begin{figure}[hbt]
    \centering
    \includegraphics[scale=0.28]{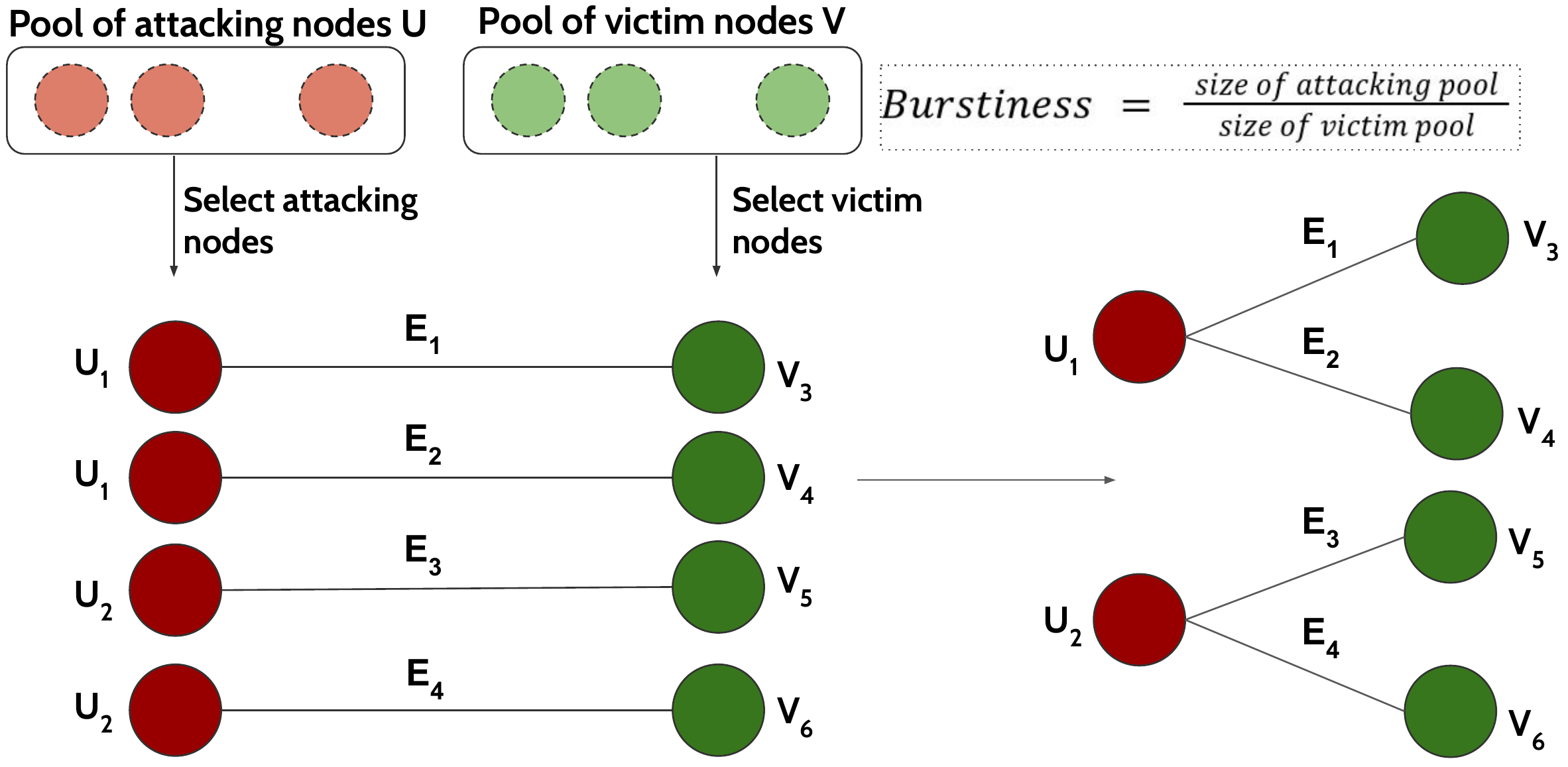} 
    \caption{Anomaly burstiness}
    \label{fig:injection}
\end{figure}

\begin{figure}[htb]
\centering
\vspace*{-6pt}
\includegraphics[scale=0.33]{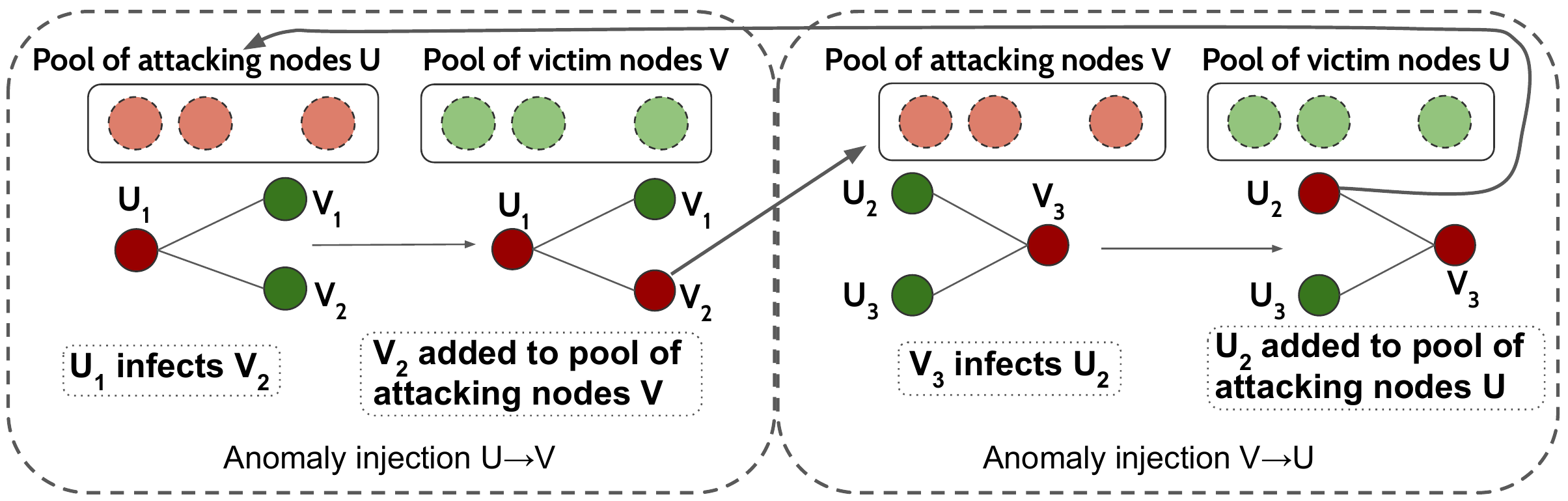} 
    \caption{Anomaly propagation}
    \vspace*{-10pt}
    \label{fig:propagation}
\end{figure}

Given the range of potential anomalies that can occur in graphs,
we define the following:
\begin{definition}
\textbf{Attacking, Victim, and Infected Node}. Attacking nodes are anomalous nodes that initiate anomalous edges with the other nodes. Victim nodes are the ones targeted by the attacking nodes. If a victim node converts to an attacking node, it becomes an infected node.
\end{definition}
\begin{definition}
\textbf{Anomaly Subgraph}. An anomaly subgraph $A_i$ is defined as a set of attacking nodes and victim nodes. $A = \{A_1, A_2, ... A_r\}$, where $r$ is the total number of injected anomalies.  
\end{definition}
\begin{definition}
\textbf{Anomaly Burstiness}. If 
an anomalous node(s) attacks a victim node(s) with a high volume of edges, burstiness can be quantified as $\frac{|\{A^{U}, A^{V}\}|}{|\{U^{victim}, V^{victim}\}|}$ where, $\{A^{U}, A^{V}\}$ represents the set of attacking nodes and $\{U^{victim}, V^{victim}\}$ represents the set of victim nodes. 
\end{definition}
\begin{definition}
\textbf{Anomaly Propagation}. Given an anomaly subgraph $A$, if there exists an edge between $U \in A$ and $V \notin A$ at time $t_1$, then there has been an event of anomaly propagation if at $t_2 > t_1$, $V \in A$.
\end{definition}
Algorithm~\ref{alg:as} presents our anomaly injection algorithm. The algorithm accepts tunable parameters such as the anomaly percentage, burstiness value, and anomaly duration. For anomalies without burstiness, we set the burstiness value to one. Additionally, we can enable propagation by setting up a propagation flag and propagation ratio. 
Either or both $U$ and $V$ nodes can be anomalous based on the flag value. For propagation, both the $U$ and $V$ nodes must be anomalous. Figures~\ref{fig:injection} and~\ref{fig:propagation} illustrate burstiness and propagation, respectively.
\begin{algorithm2e}[tbh]
\SetAlgoLined
\KwIn{Dynamic bipartite graph ($G = \{U, V, E\}$), initial number of anomalous U and V nodes ($c_{u_a}$, $c_{v_a}$), anomaly percentage ($ap$), burstiness value ($b$), propagation ratio ($p$), anomaly duration ($T_a$)}
\KwOut{dynamic bipartite graph with anomalies }
\SetKwInOut{KwIn}{Input}
\SetKwInOut{KnOut}{Output}

   $A^{U} = sampleNodes(G, c_{u_a})$\;
    $A^{V} = sampleNodes(G, c_{v_a})$\;
    $U^{victim} = sampleNodes(G, |A^{U}| * b)$\;
    $V^{victim} = sampleNodes(G, |A^{V}| * b)$\;
    
 \For{t in $T_a$}{
   $c_e = |E_t|*ap$ \;
   \If{both U and V nodes can be anomalous}{
    $add(G_t, sampleEdges(A^{U}, V^{victim}, c_e/2))$\;
    $add(G_t, sampleEdges(A^{V}, U^{victim},  c_e/2))$\;
    
     \If{Anomaly Propagation is true}{
         $add(A^{U}, sampleNodes(U^{victim}, |U^{victim}| * p))$\;
         $add(A^{V}, sampleNodes(V^{victim}, |V^{victim}| * p))$\;
     }
   }
   \If{U can be anomalous}{
    $add(G_{t},sampleEdges(A^{U}, V^{victim}, c_e))$\;
   }
    \If{V can be anomalous}{
    $add(G_{t}, sampleEdges(A^{V}, U^{victim}, c_e))$\;
   }
  }
 \caption{Anomaly Injection}
 \label{alg:as}
\end{algorithm2e}
\vspace*{-6pt}

\subsection{\textbf{Attribute generation and mapping:}}
The final step of our G2A2 methodology adds attributes to our generated dynamic bipartite graph,
using the attributes from an existing non-graph dataset $D$. 
We use two conditional tabular generative adversarial networks (CTGANs)~\cite{ctgan} to generate synthetic attributes for our graph. Based on the ground truth, we divide the original dataset $D$ into two parts: normal ($D_n$) and anomalous ($D_a$). In the absence of ground truth, we can apply clustering algorithms such as k-means with clusters equal to two. We train a CTGAN on each of $D_n$ and $D_a$ and map the generated attributes to normal and anomalous edges of our generated graph, respectively.
\setlength{\textfloatsep}{0pt}

\section{Results and Evaluation} \label{Results}

The goal of our work is to generate realistic, dynamic, attributed, bipartite graphs. However, evaluating these generated graphs remains a challenge~\cite{evaluationGG}. There exists no single metric that can precisely quantify a generated graph's ``realism'' or its level of similarity to real-world graphs. Therefore, we adopt a comprehensive three-step similarity evaluation approach that is analogous to our three-step generation methodology in order to assess our generated graph's level of similarity at the graph, anomaly, and attribute levels compared to real-world graph datasets. In our rigorous three-step evaluation approach, we compare our generated graph against three real-world graph datasets, namely P-core~\cite{pcore}, Reddit~\cite{jodie}, and Wikipedia~\cite{jodie}, based on three 
similarity criteria: \textit{(1) graph similarity}, \textit{(2) anomaly similarity}, and \textit{(3) attributes similarity}.
\subsection{Experimental Setup:}
\label{section:setup}
To quantify the similarities, we leverage past work, e.g., GraphRNN~\cite{graphrnn} and GRAN~\cite{gran}, that evaluate the performance of a model's graph generation by comparing the similarity between various graph statistics of the generated and ground truth graphs, e.g., degree distribution and clustering coefficient distribution using the maximum mean discrepancy (MMD) metric. 
The following subsections explain the  three similarity criteria and the graph statistics indicative of each similarity criteria.

\subsubsection{\textbf{Graph similarity:}}
In our graph similarity evaluation, we quantitatively compare graph statistics such as (1) the time distribution, i.e., nodes and edges distributed across time, (2) the degree distribution, and (3) the bipartite clustering coefficient (BCC) distribution~\cite{bcc} of our generated graph against the three real-world graphs using the MMD metric.
The degree distribution of a graph is a probability distribution, given by Equation~(\ref{eq:P_degree}):
\begin{equation}
\label{eq:P_degree}
            P_{degree}(k) = \frac{|\{i| degree(i) = k\}|}{N}  
\end{equation}
where $N$ is the total number of nodes.

The time distribution of a graph is a probability distribution of the number of nodes and edges over time. We compute the node time distribution of $U$ and $V$ separately. The node time and edge time distribution of a graph can be represented via Equations~(\ref{eq:pedgetime})--(\ref{eq:pnodetime}):
\begin{equation}
\label{eq:pedgetime}
            P^{edge}_{time}(t) = \frac{ |E_t|}{T} 
\end{equation}
\begin{equation}
            P^{node_U}_{time}(t) = \frac{ |U_t|}{T} 
\end{equation}
\begin{equation}
\label{eq:pnodetime}
            P^{node_V}_{time}(t) = \frac{ |V_t|}{T} 
\end{equation}
where $T$ is the total number of snapshots.

The bipartite clustering coefficient (BCC) of a graph is a measure of the local density of interactions. The BCC of a node can be calculated using the following formula: 
    \begin{equation}
            c_u = \frac{\sum_{v\in N(N(u))}^{}c_{uv}}{|N(N(u))|}  
    \end{equation}
       where $N(N(u))$ are the second-order neighbors of $u$ in $G$ excluding $u$, and $c_{uv}$ is the pairwise clustering coefficient between nodes $u$ and $v$.
       For our experiments, we define $c_{uv}$ using the following equation:
       \begin{equation}
            c_{uv} = \frac{|N(u) \cap N(v)|}{|N(u) \cup N(v)|}
        \end{equation}
        We calculate the average bipartite clustering coefficient as:
        \begin{equation}
            BCC_X = \frac{1}{|X|}\sum_{v\in X}^{} c_v
        \end{equation}
            where $X \in \{U,V\}$. 

\subsubsection{\textbf{Anomaly similarity:}}
We quantify anomaly similarity separately from graph similarity since anomalies are rare occurrences and typically do not affect the overall graph statistics. 
To compare the anomalies, we filter the anomaly subgraphs from both the generated graph and the real-world graphs. Similar to graph similarity, we then compare the BCC distribution of the two anomaly subgraphs using the MMD metric.

\subsubsection{\textbf{Attributes similarity:}}
To quantify the similarity of attributes between our generated and real-world graphs, we check if both sets of attributes have been drawn from the same distribution and use the MMD metric to verify the distribution similarity~\cite{twosample}.
To do so, we had to preprocess the data by normalizing it and then create a histogram with a specific bin size. In this paper, we fixed the bin size to 100.

\subsection{Real-world Graph Datasets}
For our evaluation,
we use
three real-world datasets: P-core, Reddit, and Wikipedia. P-core \cite{pcore} is a dataset provided by 
Advanced Research Computing (ARC) at Virginia Tech (VT). Reddit and Wikipedia
are from the JODIE~\cite{jodie} repository. Table~\ref{table: dataset} provides a quantitative summary of the datasets. 
\vspace*{-5pt}
\begin{itemize}
\itemsep0em
\item \textbf{P-core:} This dataset contains network traffic flows (E) information going through the edge server from the VT domain ($U$) to the rest-of-the-world (RoW) domain ($V$) and vice versa. 
The anomaly ($A^e$) for this dataset is a staged Mirai-botnet attack.
\item \textbf{Reddit:} This dataset contains posts ($E$) made by users ($U$) on subreddits ($V$) in a month ($T$). 
Additionally, anomaly labels ($A^e$) at the edge level represent the interactions after which the user got banned. 
\item \textbf{Wikipedia:} This dataset contains edits ($E$) done by users ($U$) on Wikipedia pages ($V$) recorded over a month. 
Additionally, anomaly labels ($A^e$) at the edge level represent the last edit made by the user before they got banned. 
\end{itemize}

\vspace*{-10pt}
\begin{table}[tbh]
\centering
\caption{Summary of real-world graph datasets}
\scalebox{0.72}{
\renewcommand{\arraystretch}{1}
\begin{tabular}{|c|c|c|c|c|c|c|}
\hline
\textbf{Dataset}       & \boldmath{$|U|$}       & \boldmath{$|V|$}    & \boldmath{$|E|$}       & \boldmath{$|A^e|$}       & \boldmath{$|F|$}      & \boldmath{$T (hours)$}       \\ \hline \hline
P-core                     & 157225      & 96037   & 597098      & 6012 (1\%)        &  14        & 48           \\ \hline
Reddit                    & 10000       & 984      & 672447      & 366 (0.05\%)      &  172       & 744         \\ \hline
Wikipedia                 & 8227        & 1000     & 157474      & 217 (0.14\%)      &  172       & 744         \\ \hline

\end{tabular}
}
\label{table: dataset}
\end{table} 
\vspace*{-10pt}
    \begin{figure*}[tbh]\centering
     \centering
     \begin{subfigure}[b]{0.22\textwidth}
         \centering
         \includegraphics[width=1.1\textwidth]{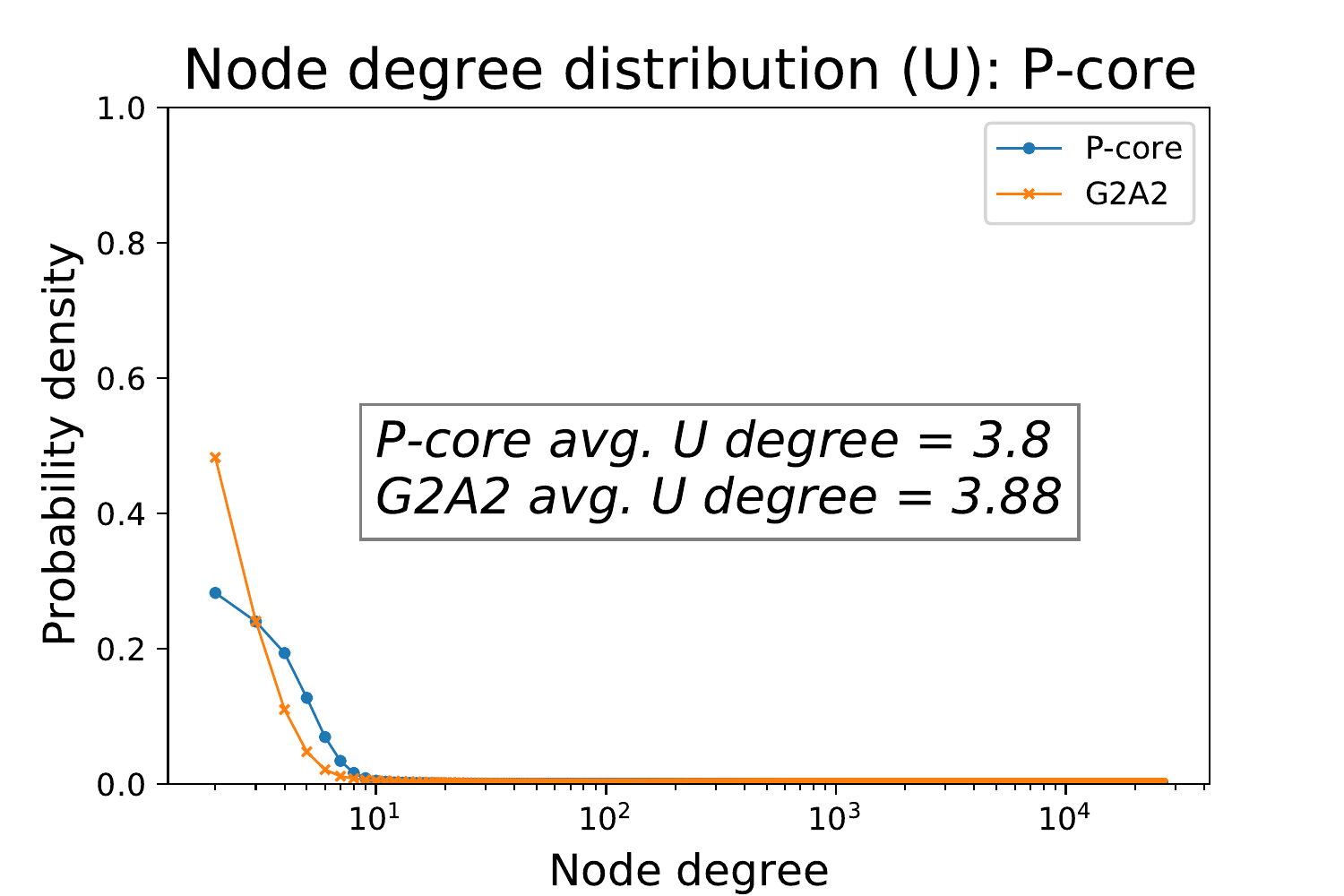}
         \caption{VT IPs }
         \label{fig:pcore_node_1}
     \end{subfigure}
     \hfill
     \begin{subfigure}[b]{0.22\textwidth}
         \centering
         \includegraphics[width=1.1\textwidth]{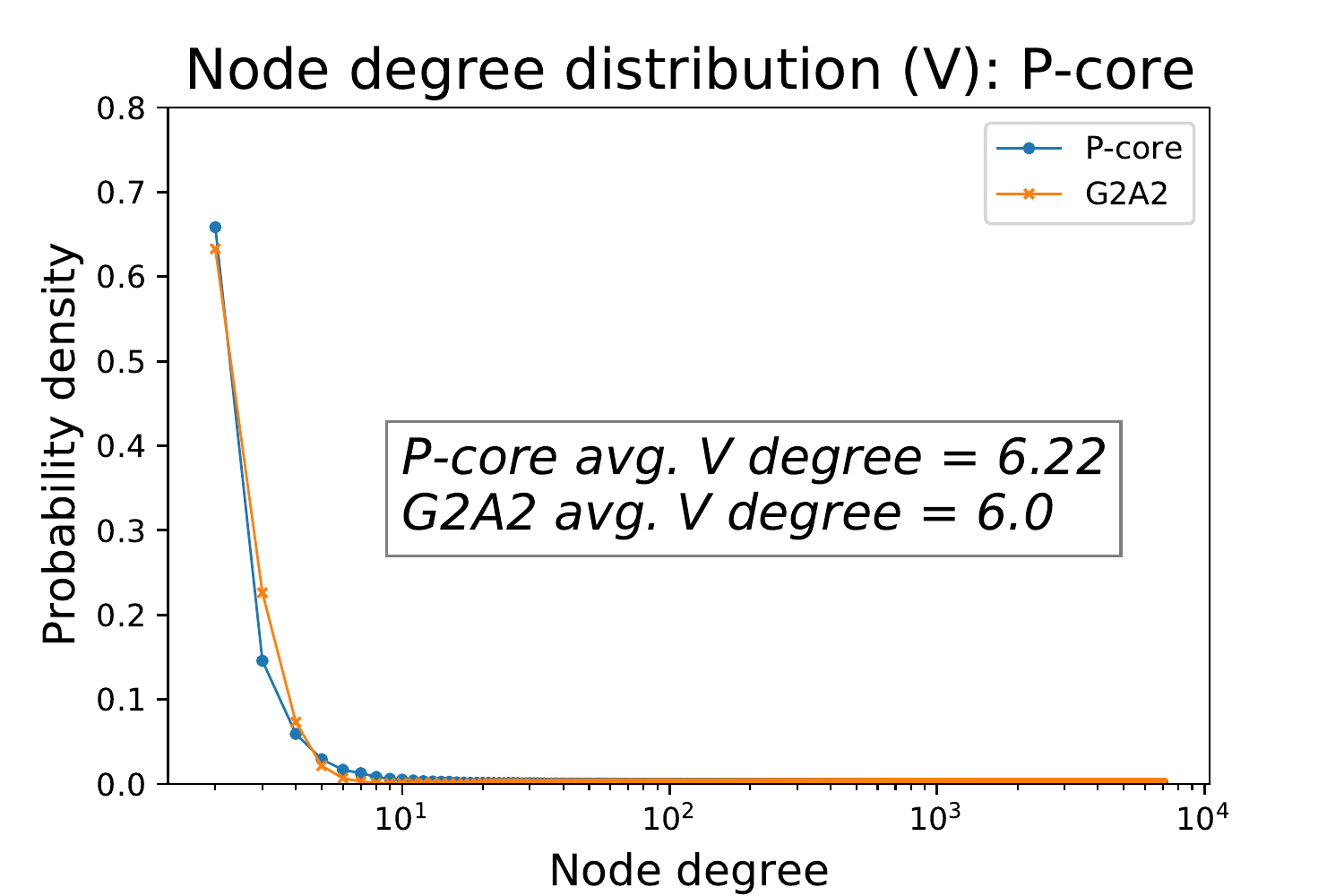}
         \caption{ROW IPs }
         \label{fig:pcore_node_2}
     \end{subfigure}
     \hfill
     \begin{subfigure}[b]{0.22\textwidth}
         \centering
         \includegraphics[width=1.1\textwidth]{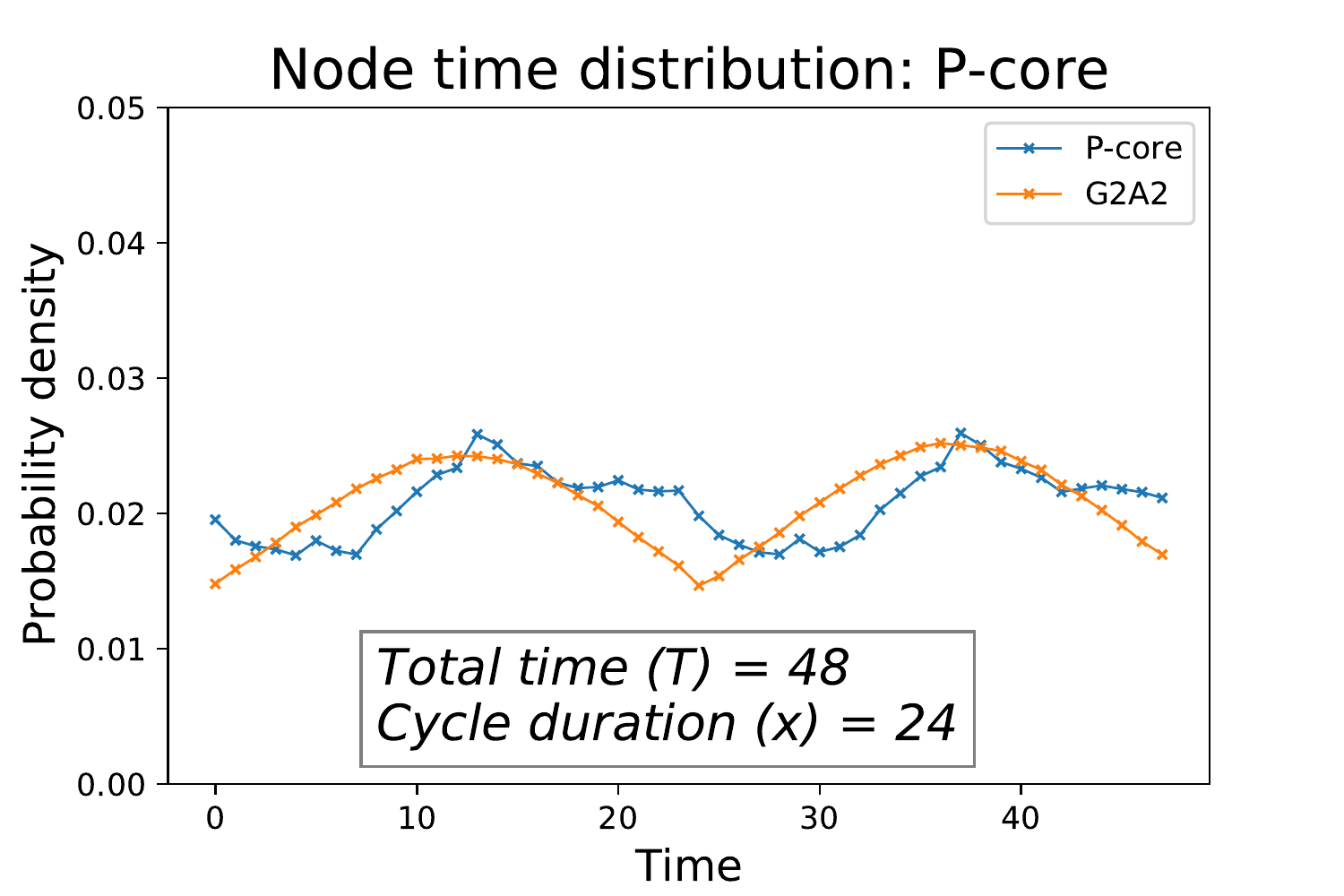}
         \caption{Node time distribution}
         \label{fig:pcore_node_time}
     \end{subfigure}
     \hfill
     \begin{subfigure}[b]{0.22\textwidth}
         \centering
         \includegraphics[width=1.1\textwidth]{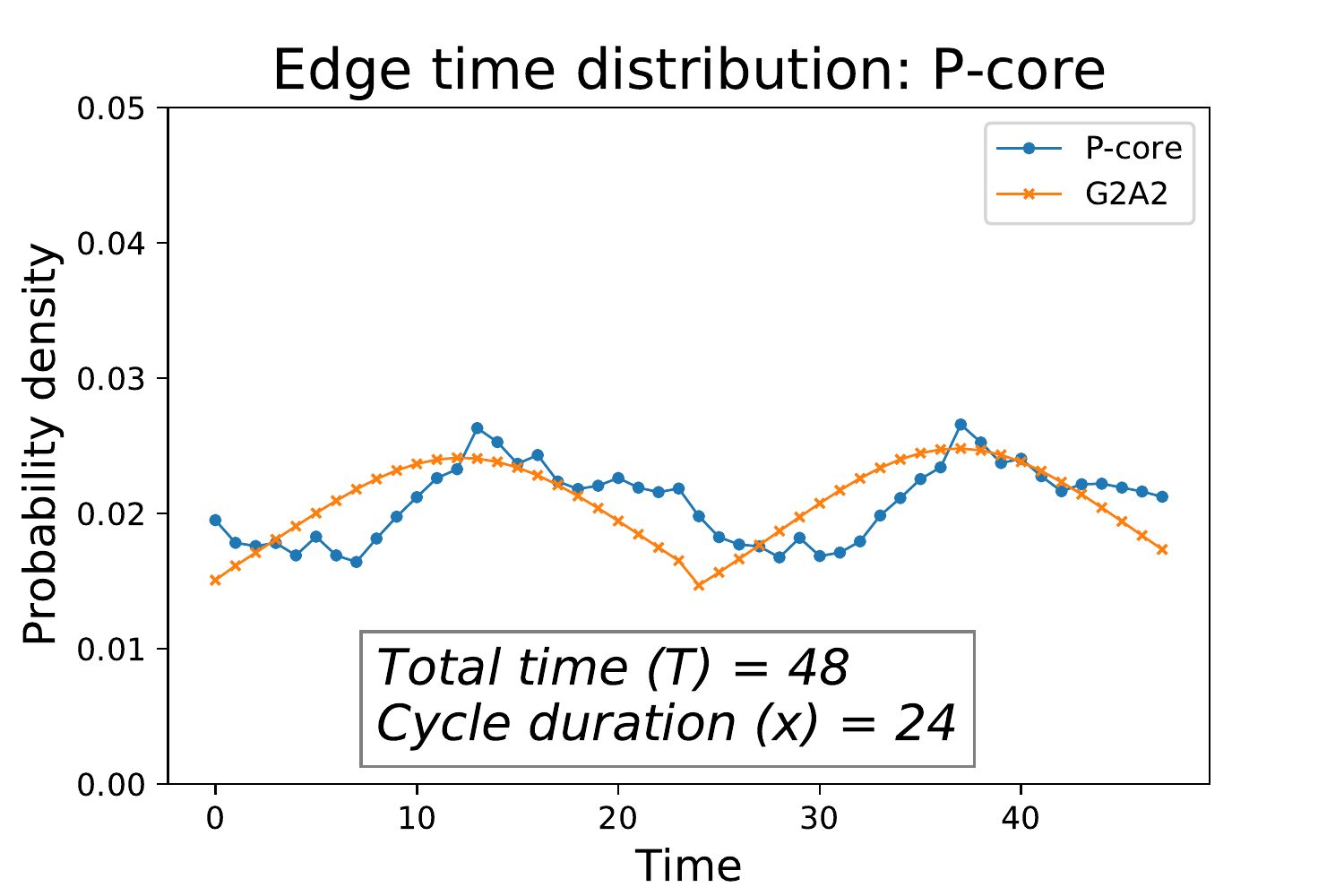}
         \caption{Edge time distribution}
         \label{fig:pcore_time}
     \end{subfigure}
     \hfill
     \begin{subfigure}[b]{0.22\textwidth}
         \centering
         \includegraphics[width=1.1\textwidth]{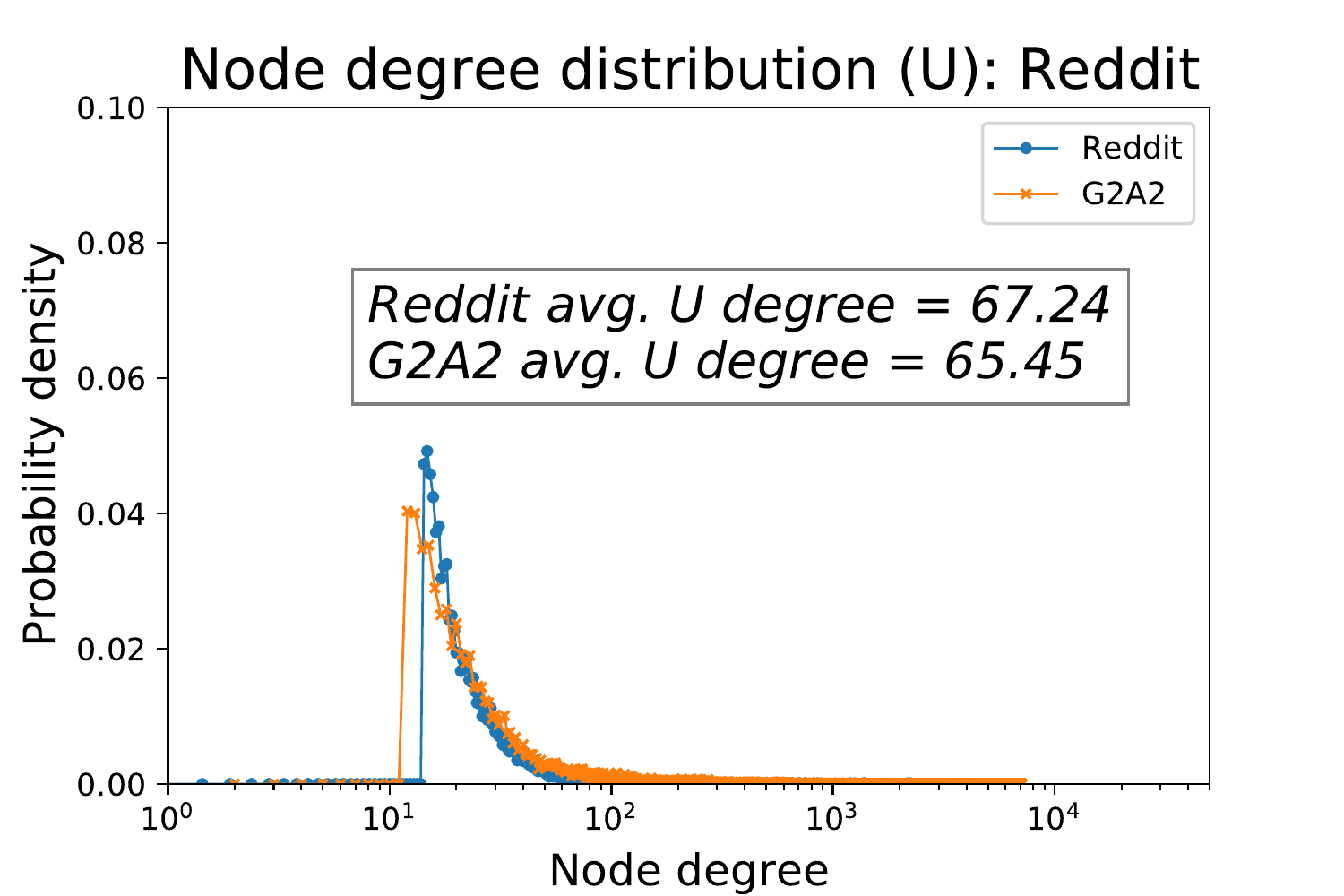}
         \caption{Users}
         \label{fig: reddit_node_1}
     \end{subfigure}
     \hfill
     \begin{subfigure}[b]{0.22\textwidth}
         \centering
         \includegraphics[width=1.1\textwidth]{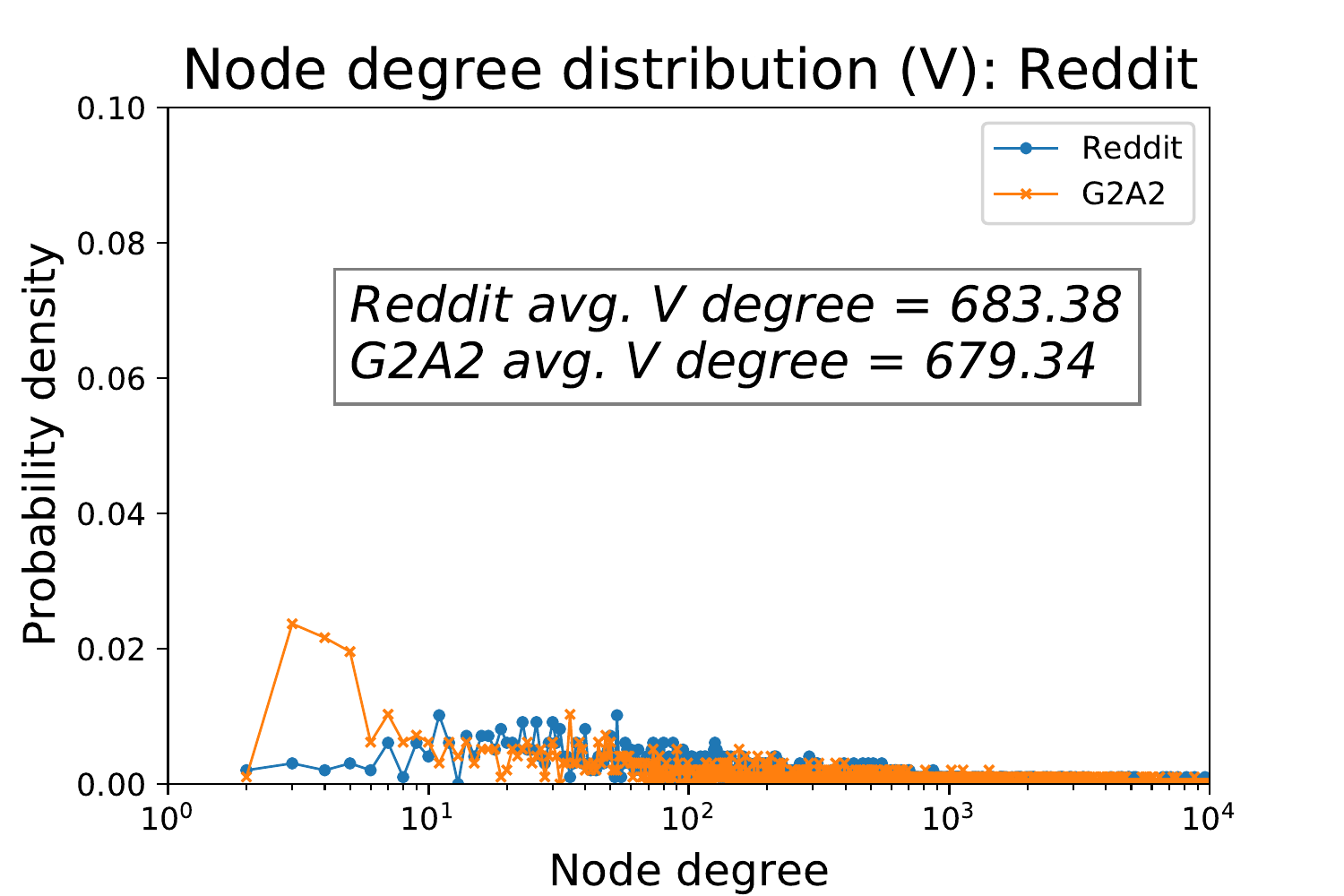}
         \caption{Subreddit}
         \label{fig:reddit_node_2}
     \end{subfigure}
     \hfill
     \begin{subfigure}[b]{0.22\textwidth}
         \centering
         \includegraphics[width=1.0\textwidth]{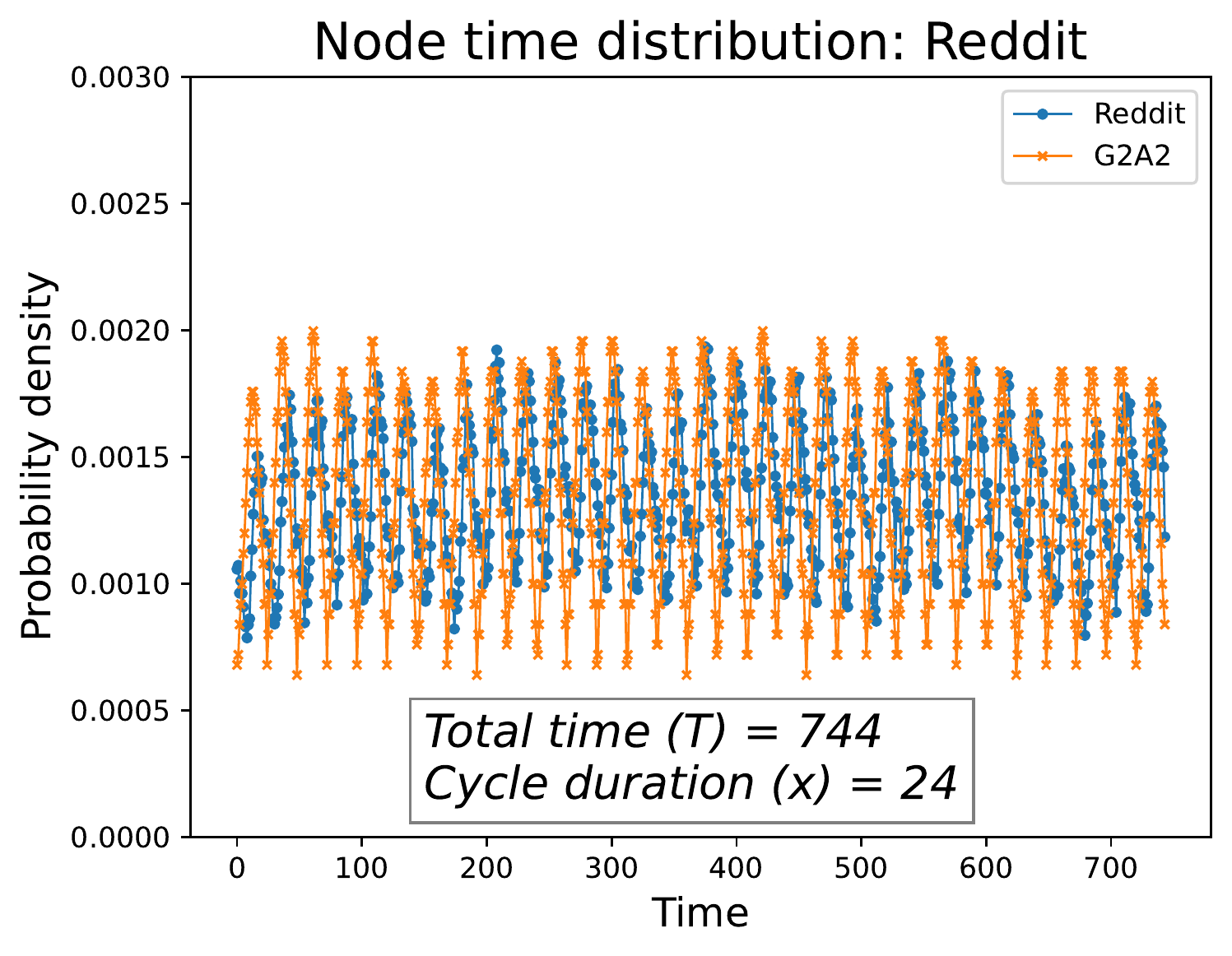}
         \caption{Node time distribution}
         \label{fig:reddit_node_time}
     \end{subfigure}
     \hfill
     \begin{subfigure}[b]{0.22\textwidth}
         \centering
         \includegraphics[width=1.0\textwidth]{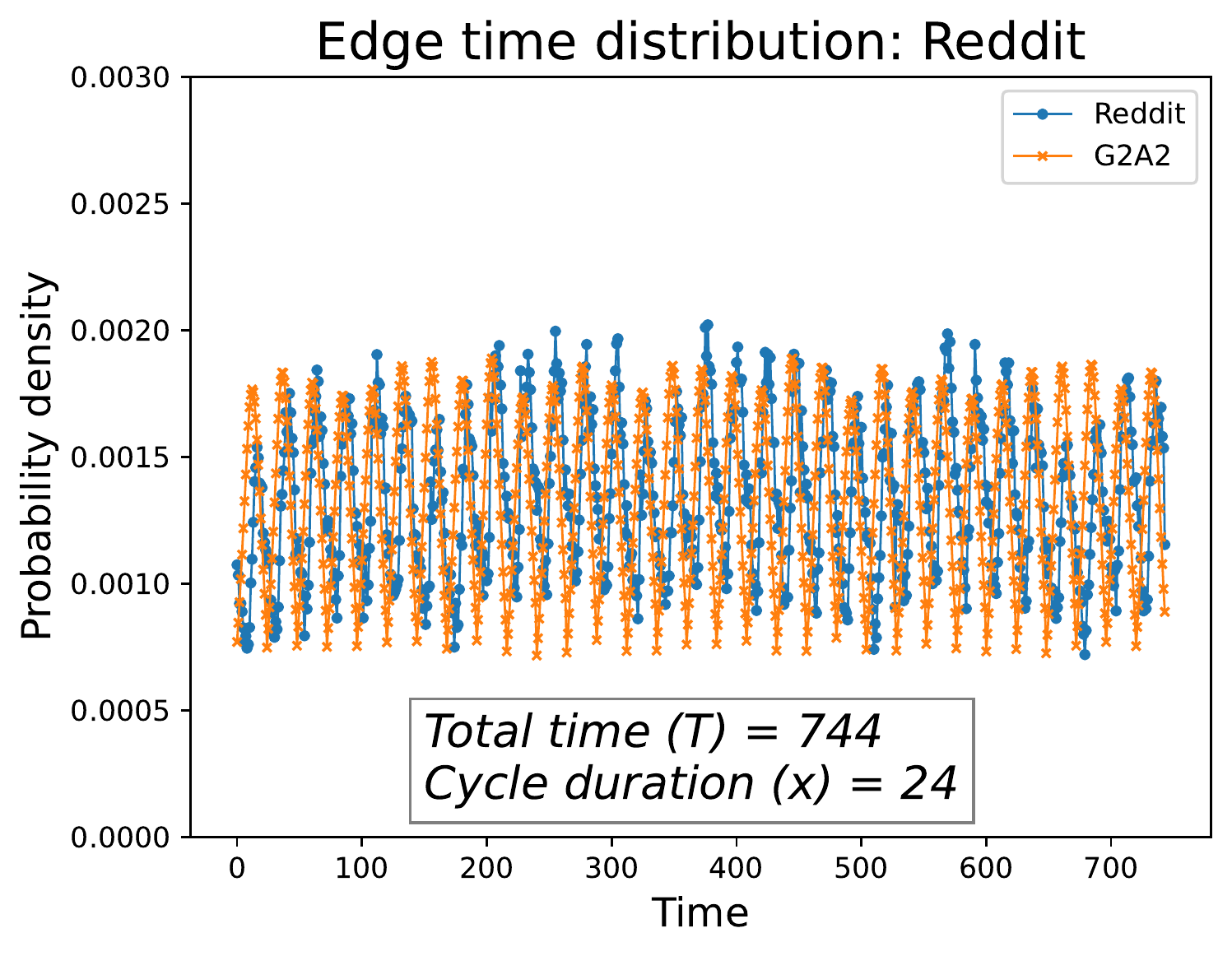}
         \caption{Edge time distribution}
         \label{fig:reddit_edge_time}
     \end{subfigure}
     \begin{subfigure}[b]{0.22\textwidth}
         \centering
         \includegraphics[width=1.1\textwidth]{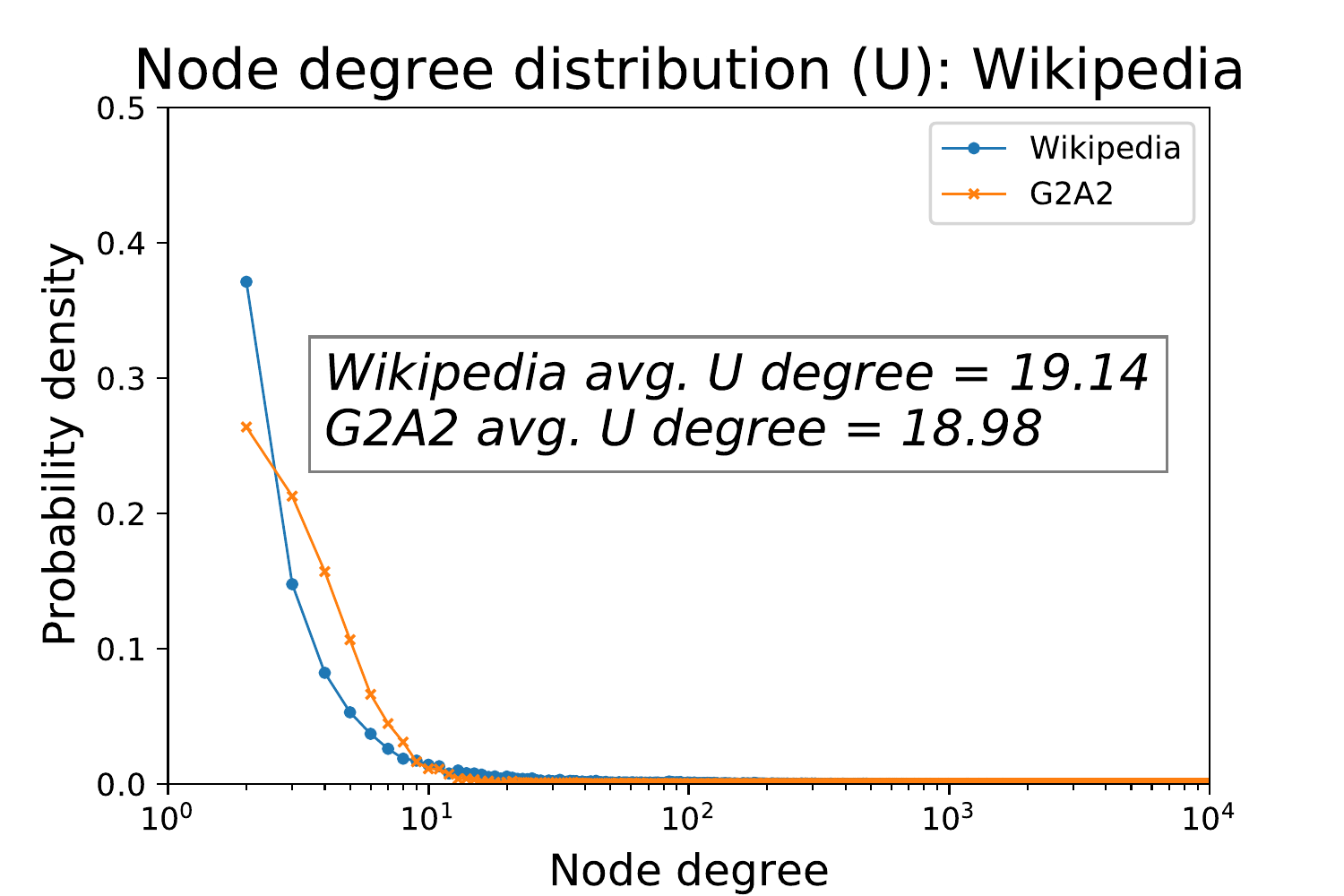}
         \caption{Users}
         \label{fig:wiki_node_1}
     \end{subfigure}
     \hfill
     \begin{subfigure}[b]{0.22\textwidth}
         \centering
         \includegraphics[width=1.1\textwidth]{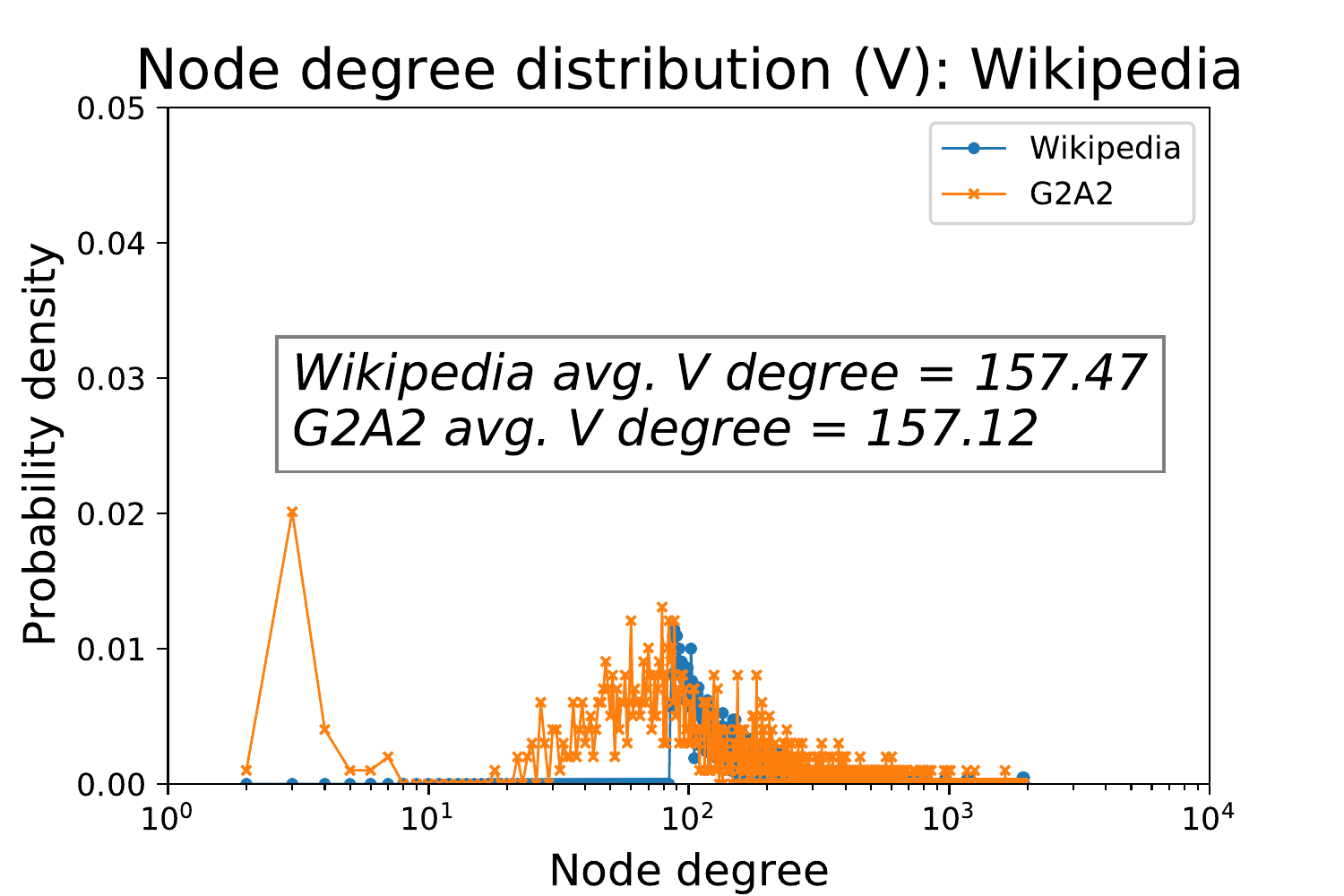}
        \caption{Pages}
        \label{fig:wiki_node_2}
     \end{subfigure}
     \hfill
     \begin{subfigure}[b]{0.22\textwidth}
         \centering
         \includegraphics[width=1.1\textwidth]{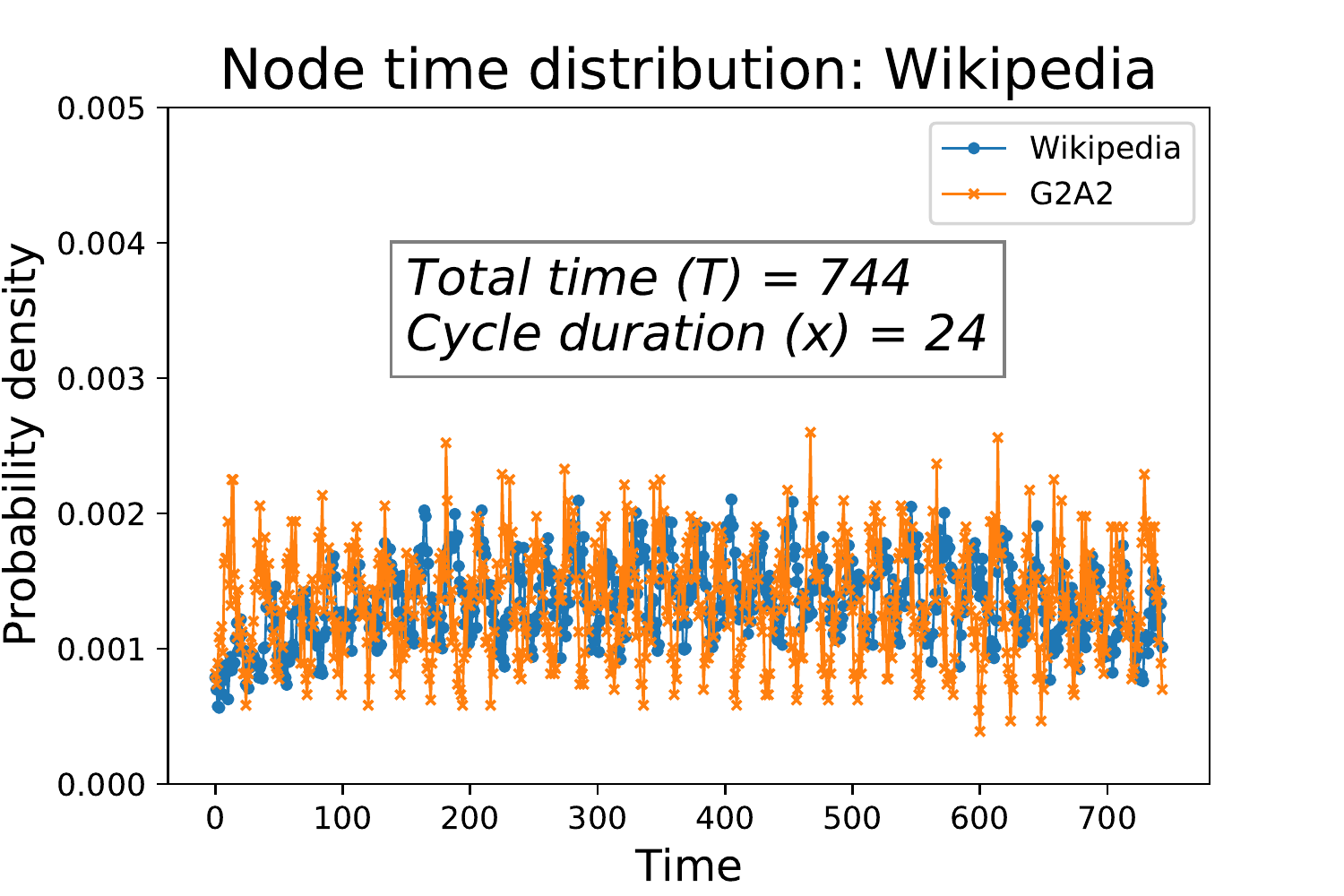}
         \caption{Node time distribution}
         \label{fig:wiki_node_time}
     \end{subfigure}
     \hfill
     \begin{subfigure}[b]{0.22\textwidth}
         \centering
         \includegraphics[width=1.1\textwidth]{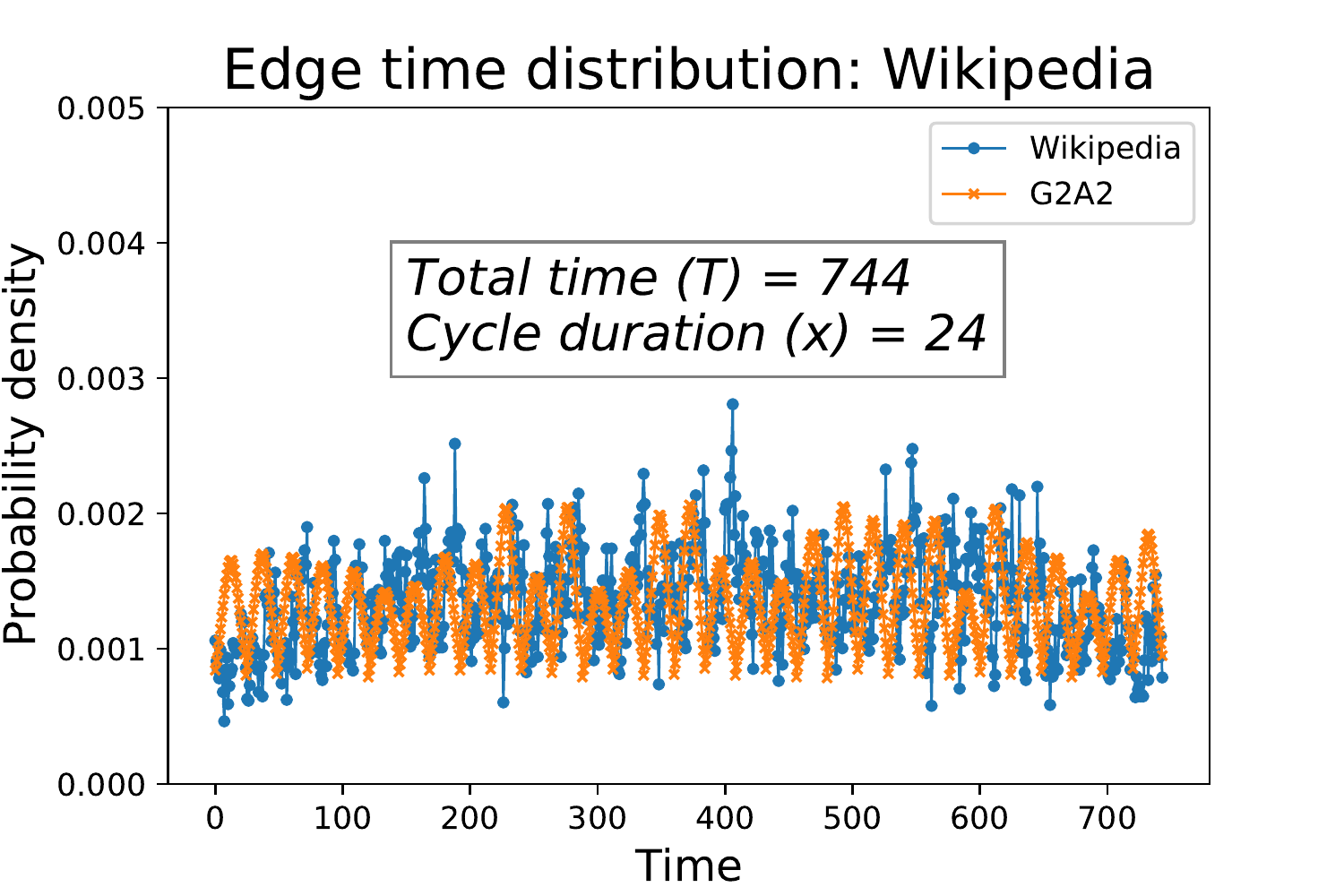}
         \caption{Edge time distribution}
         \label{fig:wiki_time}
     \end{subfigure}
        \caption{G2A2-generated vs.\ real-world graphs: P-core (a, b, c, d), Reddit (e, f, g, h) and Wikipedia (i, j, k, l)} 
        \label{fig:node and degree distribution}
\end{figure*}
\subsection{Generating Realistic Graphs:}

By setting the parameters in G2A2 appropriately, we can generate 
realistic graphs that are similar to real-world graphs with respect to graph similarity, anomaly similarity, and attribute similarity. In addition, we compare G2A2 to other state-of-the-art methodologies to show that G2A2 outperforms them by reducing the MMD metric over various graph distributions and computation time.

\subsubsection{\textbf{Setting Cauchy and gamma parameters:}}
\begin{figure*}[tbh]\centering
     \centering
     \begin{subfigure}[b]{0.17\textwidth}
         \centering
         \includegraphics[width=1.2\textwidth]{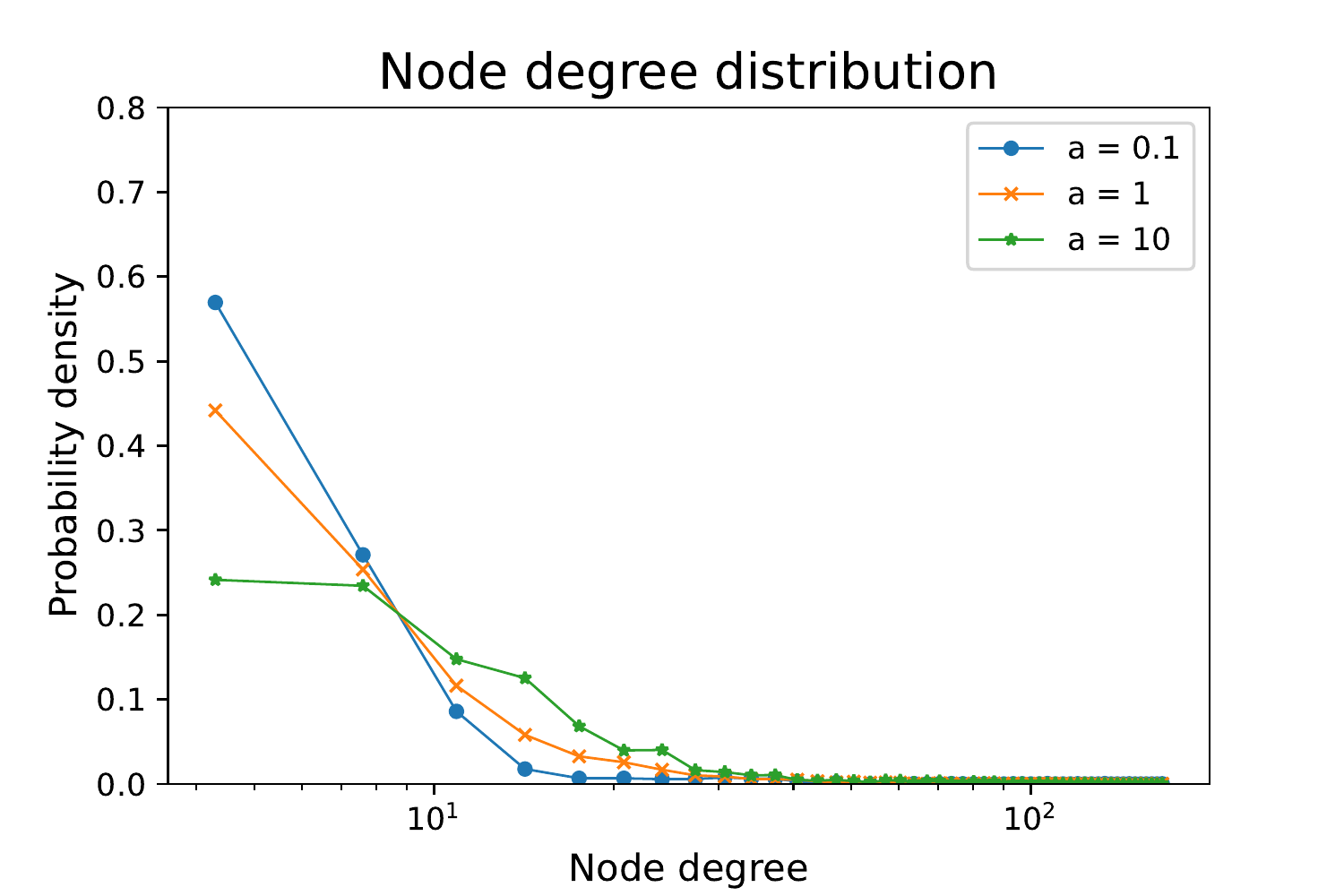}
         \caption{Gamma: a}
         \label{fig:node_a}
     \end{subfigure}
     \hfill
     \begin{subfigure}[b]{0.17\textwidth}
         \centering
         \includegraphics[width=1.2\textwidth]{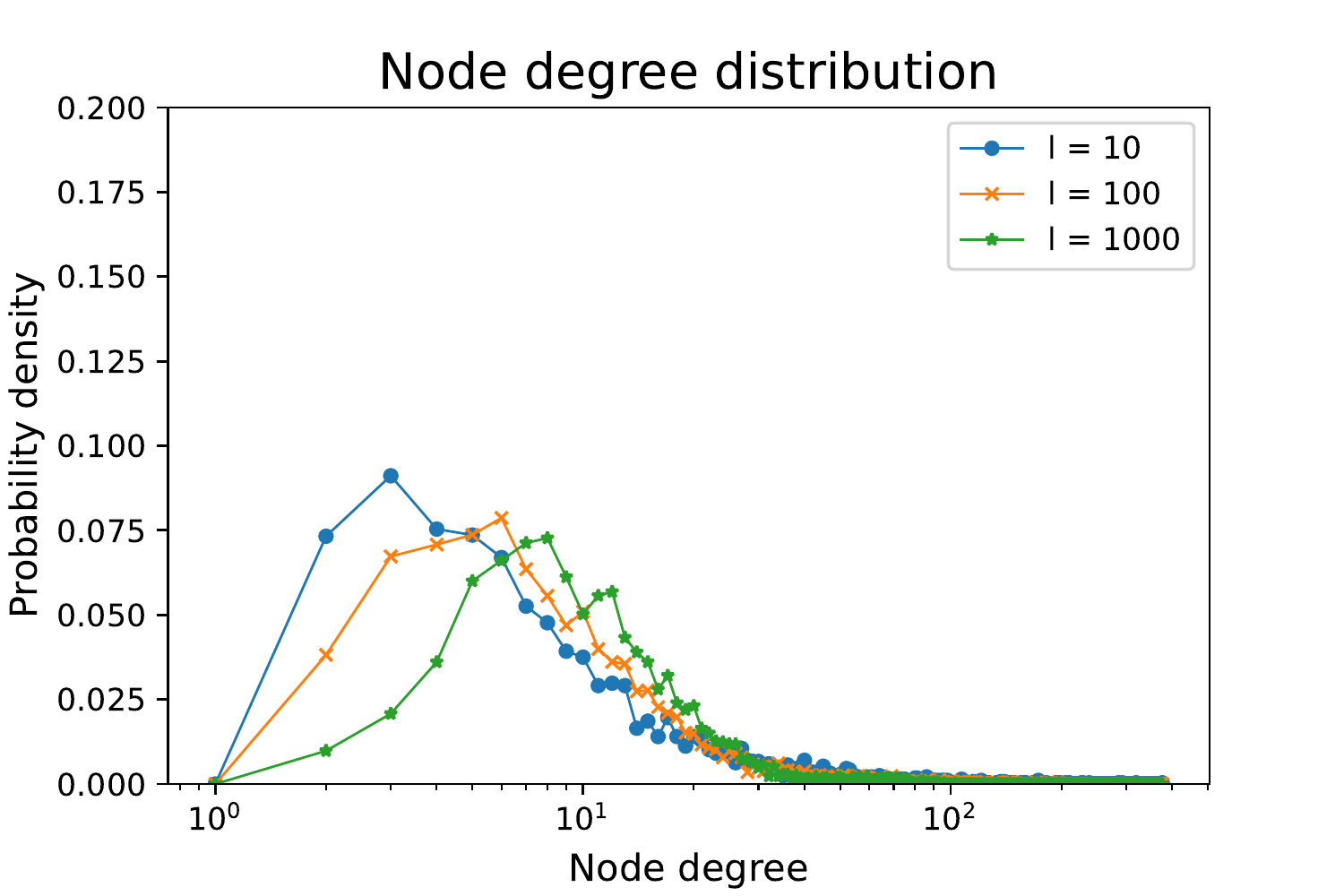}
         \caption{Gamma: l}
         \label{fig:node_l}
     \end{subfigure}
     \hfill
     \begin{subfigure}[b]{0.17\textwidth}
         \centering
         \includegraphics[width=1.2\textwidth]{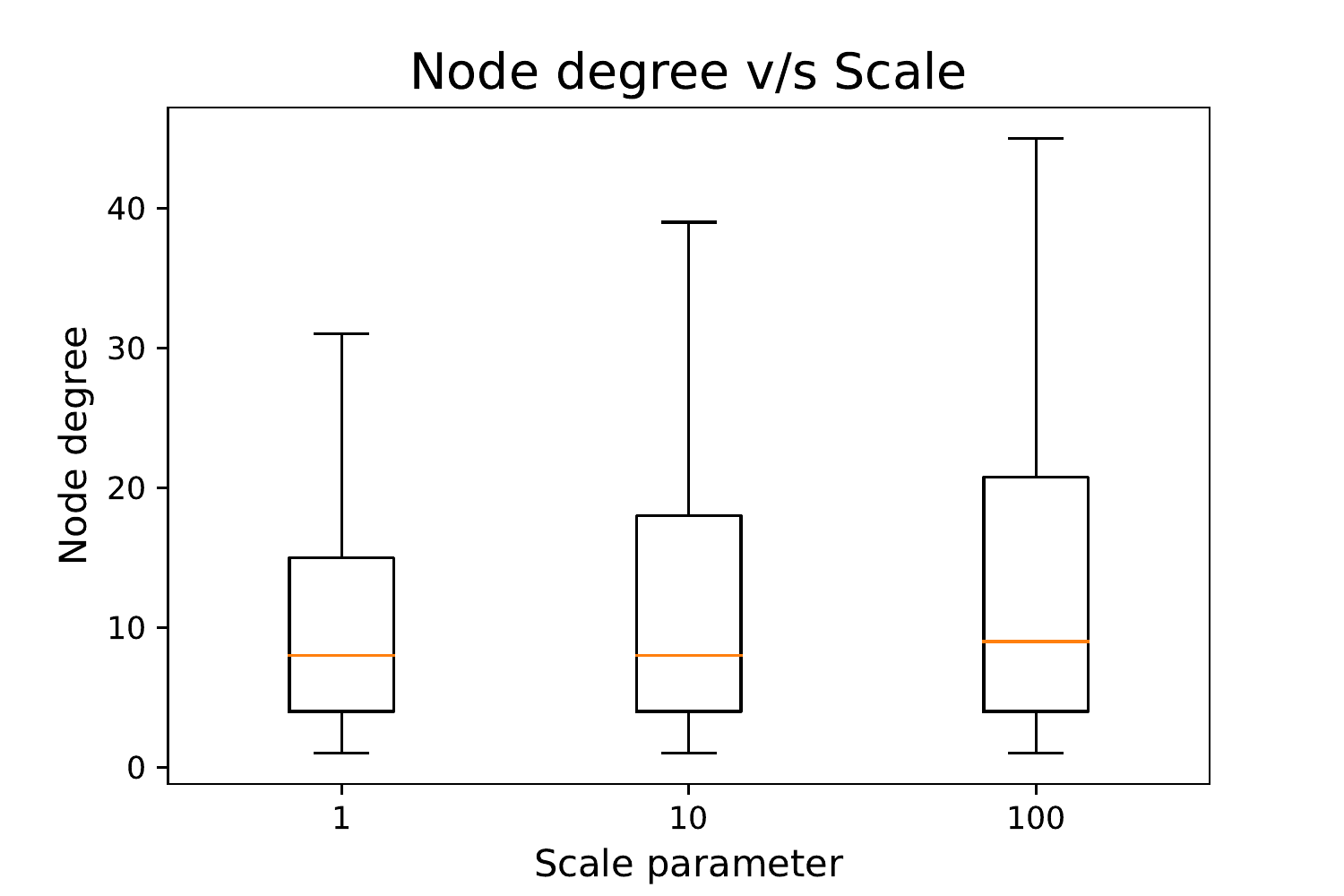}
         \caption{Gamma: s}
         \label{fig:node_s}
     \end{subfigure}
     \hfill
     \begin{subfigure}[b]{0.17\textwidth}
         \centering
         \includegraphics[width=1.2\textwidth]{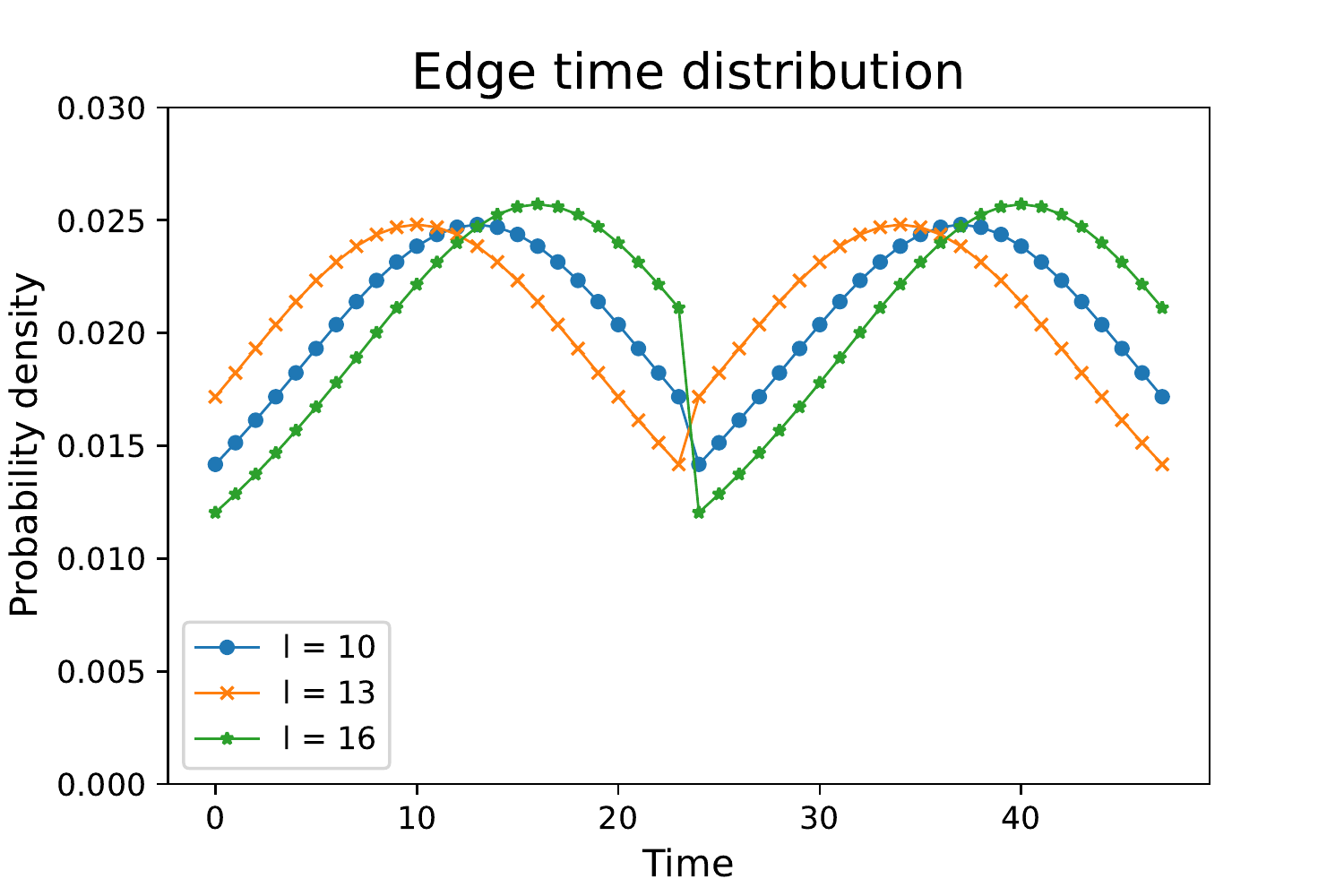}
         \caption{Cauchy: l}
         \label{fig:time_l}
     \end{subfigure}
     \hfill
     \begin{subfigure}[b]{0.17\textwidth}
         \centering
         \includegraphics[width=1.2\textwidth]{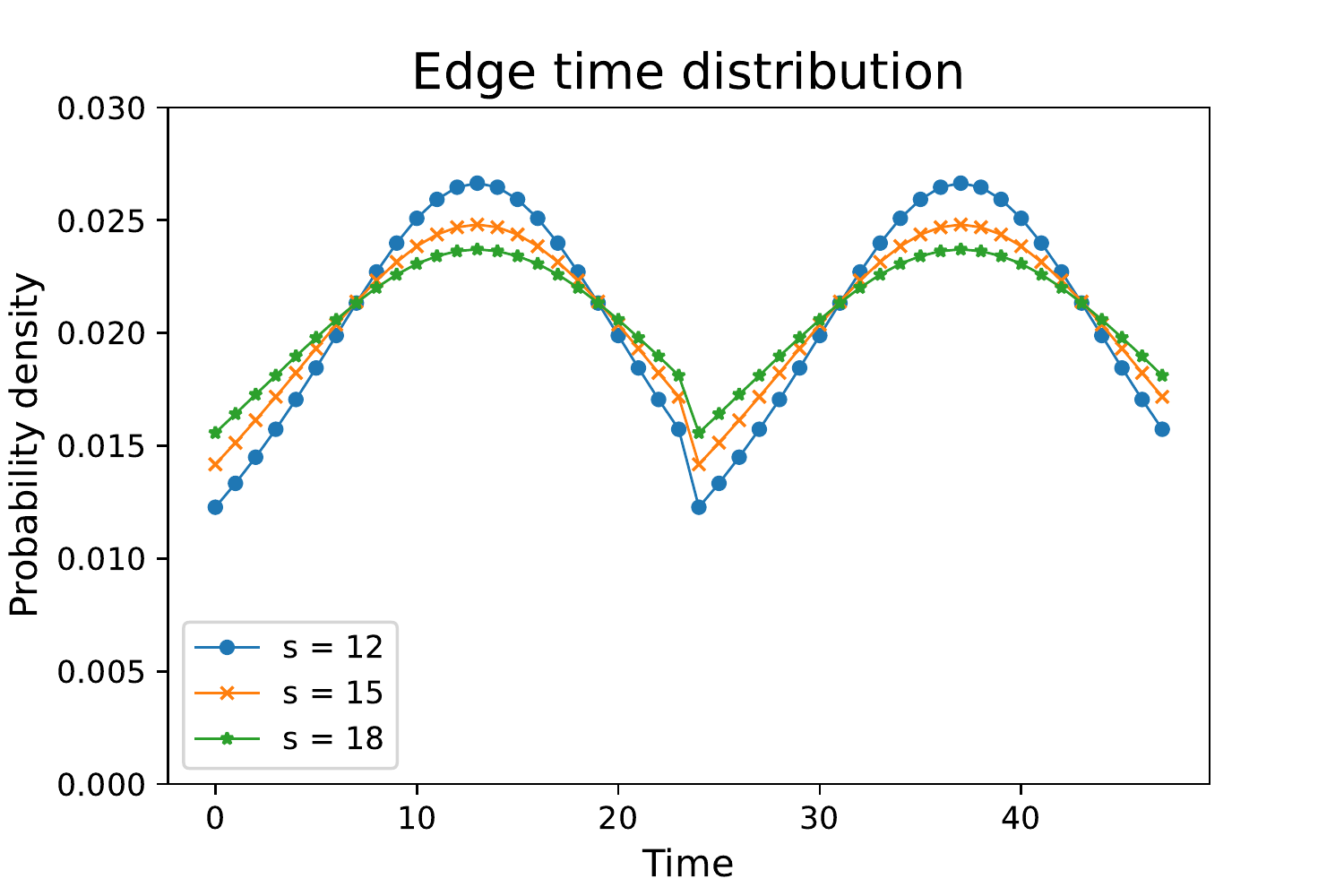}
         \caption{Cauchy: s}
         \label{fig:time_s}
     \end{subfigure}
        \caption{Parameter analysis of gamma (a, b, c)  on the node degree, and  Cauchy (d, e) on edge time distribution.}
        \vspace*{-10pt}
        \label{fig:parameter study}
\end{figure*}

While determining the exact values of the Cauchy and gamma distribution parameters to generate a realistic graph may be difficult, it is easier to estimate them based on their significance in shaping the overall desired distribution (a distribution similar to that of a real-world graph).  
For instance, the Cauchy distribution parameter location ($l$) and scale ($s$) determine the position of the peak and length of peak to trough, respectively. 
Similarly, for the gamma distribution, we have three parameters: shape ($a$), location ($l$), and scale ($s$). The shape ($a$) determines the skewness of the degree distribution. The lower the shape parameter, the more skewed the graph is. The location parameter ($l$) specifies the minimum probability of the smallest node degree. The higher the location parameter, the greater the minimum node degree. The gamma distribution's scale parameter (s) determines the difference between the lowest degree and highest degree. The higher the scale parameter, the more significant the difference. Figure~\ref{fig:parameter study} shows the effect of setting the gamma and Cauchy function parameters.
\begin{table*}[tbh]
\caption{Comparison of G2A2 with the other graph generation models. For all MMD results, \textbf{lower is better}.}
\scalebox{0.60}{
\renewcommand{\arraystretch}{1}
\begin{tabular}{|c|c|c|c|c|c|c|c|c|c|c|c|c|c|c|c|}
\hline
\multirow{3}{*}{\textbf{Methods}} & \multicolumn{5}{c|}{\textbf{P-core}}                                     & \multicolumn{5}{c|}{\textbf{Reddit}}                                     & \multicolumn{5}{c|}{\textbf{Wikipedia}}                                  \\ \cline{2-16} 
                         & \multicolumn{3}{c|}{\textbf{Graph Similarity}} & \textbf{Anomaly}    & \textbf{Attribute}  & \multicolumn{3}{c|}{\textbf{Graph Similarity}} & \textbf{Anomaly}    & \textbf{Attribute}  & \multicolumn{3}{c|}{\textbf{Graph Similarity}} & \textbf{Anomaly}    & \textbf{Attribute}  \\ \cline{2-4} \cline{7-9} \cline{12-14}
                         & \textbf{Degree}      & \textbf{BCC}        & \textbf{Time}       & \textbf{Similarity} & \textbf{Similarity} & \textbf{Degree}      & \textbf{BCC}        & \textbf{Time}       & \textbf{Similarity} & \textbf{Similarity} & \textbf{Degree}      & \textbf{BCC}        & \textbf{Time}       & \textbf{Similarity} & \textbf{Similarity} \\ \hline
E-R                      & 1.047       & 0.957      & NA         & NA         & NA         & 0.750        & 0.966      & NA         & NA         & NA         & 0.670        & 0.911      & NA         & NA      & NA         \\ \hline
Kronecker                & 0.578       & 0.352      & 0.789      & NA         & NA         & 0.539       & 0.347      & 0.616      & NA         & NA         & 0.511       & 0.398      & 0.675      & NA         & NA         \\ \hline
Kronecker + attributes   & 0.578       & 0.352      & 0.789      & NA         & 0.584      & 0.539       & 0.347      & 0.616      & NA         & 0.712         & 0.511       & 0.398      & 0.675      & NA         & 0.649      \\ \hline
MAG                      & 1.778       & 0.544      & NA         & NA         & 1.047      & 1.686       & 0.541      & NA         & NA         & 1.854      & 1.725       & 0.522      & NA         & NA         & 1.959      \\ \hline
\textbf{G2A2}                     & \textbf{0.128}       & \textbf{0.194}      & \textbf{0.003}      & \textbf{0.064}      & \textbf{0.076}      & \textbf{0.142}       & \textbf{0.252}      & \textbf{0.016}      & \textbf{0.165}      & \textbf{0.044}      & \textbf{0.150}      & \textbf{0.210}      & \textbf{0.020}      & \textbf{0.255}      & \textbf{0.010}      \\ \hline
\end{tabular}
}
\vspace*{-10pt}
\label{table: graph evaluation}
\end{table*}

\subsubsection{\textbf{G2A2-generated graphs vs.\ real-world graphs:}}
Our G2A2 methodology generates graphs similar to the following three real-world graphs: P-core, Reddit, and Wikipedia. We evaluate their quality by plotting and comparing the degree and time distribution (node + edge) of the generated graph versus the real graph, as shown in Figure~\ref{fig:node and degree distribution}. Parameters used for the generation have been obtained heuristically, as explained in the previous section and do not represent the optimal value. The objective is to demonstrate the capability of G2A2 in generating realistic graphs even without the optimal values of parameters in place. 
An important note is that when we generate graphs for the social media graphs (Reddit) and article graphs (Wikipedia), the propagation flag for the anomaly injection is false. That means the victim nodes do not participate in the attack. In contrast, the propagation flag is true when generating graphs for the internet traffic graph (P-core). That means the victim nodes participate in the attack if they are infected.

\subsubsection{\textbf{G2A2 vs.\ the other models:}}
We compare the quality of the generated graphs via G2A2 against the ones generated by existing models such as Erdos-Renyi (E-R)~\cite{ermodel}, Kronecker~\cite{kroneckergraph}, and Multiplicative Attribute Graph (MAG)~\cite{mag}. Also, we have included Kronecker with randomly generated attributes as a baseline to show how our model compares to randomly generated attributes. However, all of the models could only generate parts of a dynamic attributed bipartite graph with instances of anomalies. Therefore we could only compare the relative parts of the graph these models can generate with G2A2-generated graphs. 
Kronecker's performance is the best of all the other models that we compared G2A2 with. The Kronecker graph generator uses an initiator matrix with a given pattern and can generate bipartite graphs. We obtain the initiator matrix using Kroneckerfit (provided by SNAP library~\cite{snap}) that can learn parameters for the model. Kronecker can also generate realistic static and dynamic properties of a graph. However, the generator does \emph{not} provide a timestamp to the edges; we had to equally divide the edges into different timestamps to evaluate the temporal component of the graph. 

We also evaluated G2A2 versus a modified version of Kronecker by generating a random matrix with the attribute dimension's size and mapping it to the edges of the Kronecker graph. As shown in Table~\ref{table: graph evaluation},  G2A2 delivers the best results, followed by Kronecker+attributes.

\begin{figure*}[tbh]\centering
 \centering
 \begin{subfigure}[b]{0.32\textwidth}
     \centering
     \includegraphics[scale=0.26]{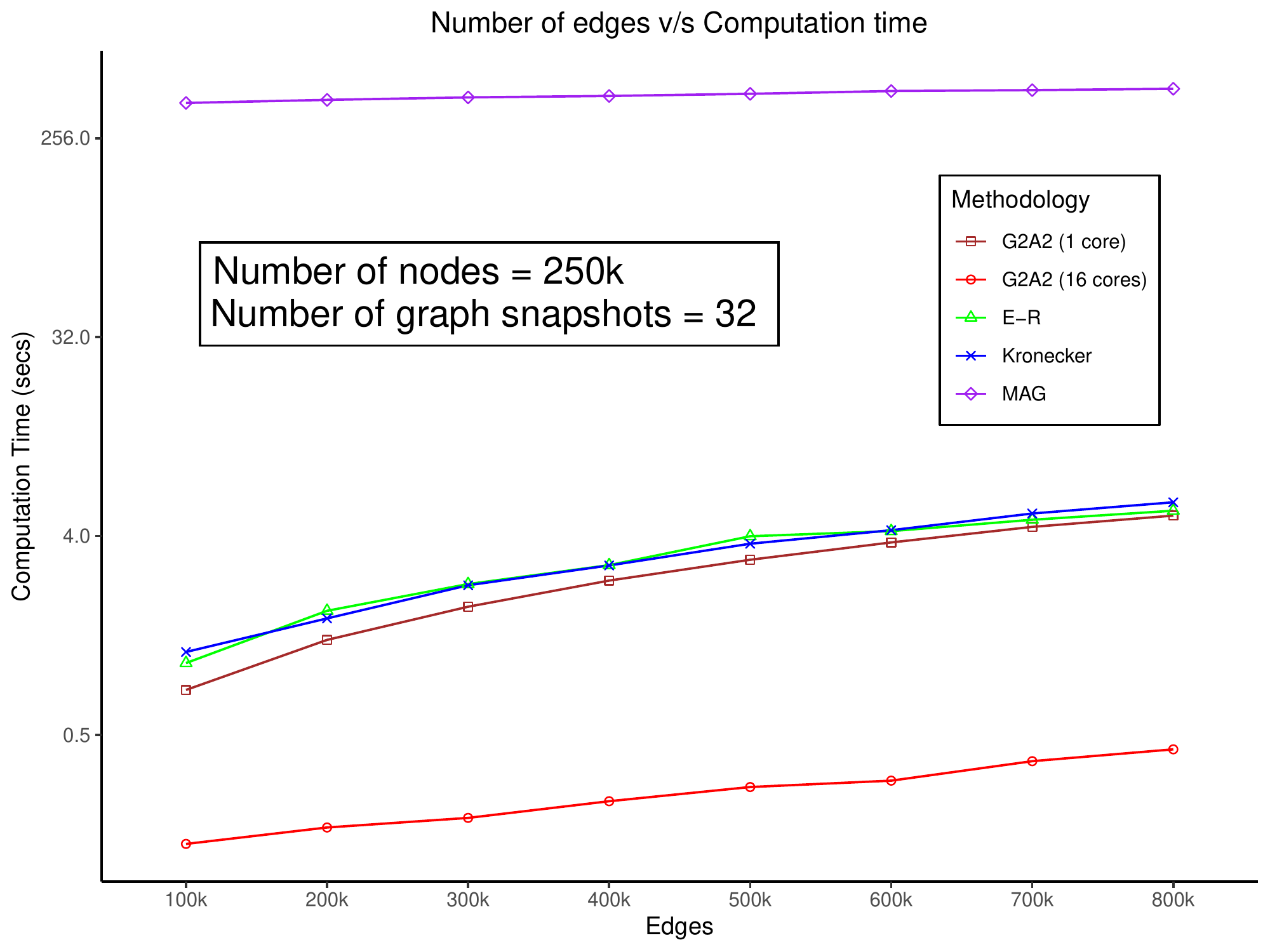}
     \caption{Edge analysis}
     \label{fig:EdgeVsTime}
 \end{subfigure}
 \hfill
 \begin{subfigure}[b]{0.32\textwidth}
     \centering
     \includegraphics[scale=0.26]{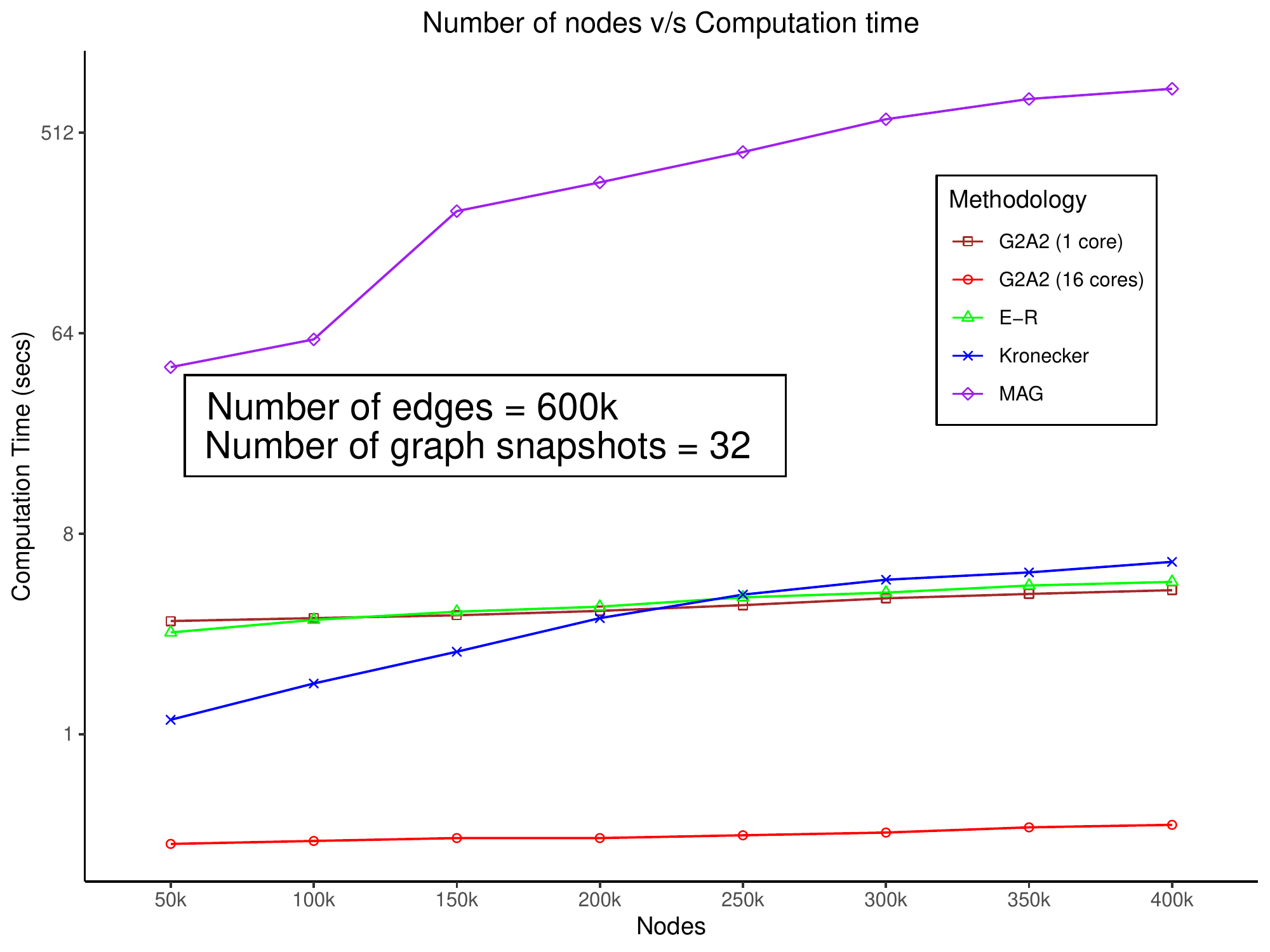}
     \caption{Node analysis}
     \label{fig:NodesVsTime}
 \end{subfigure}
 \hfill
 \begin{subfigure}[b]{0.32\textwidth}
     \centering
     \includegraphics[scale=0.26]{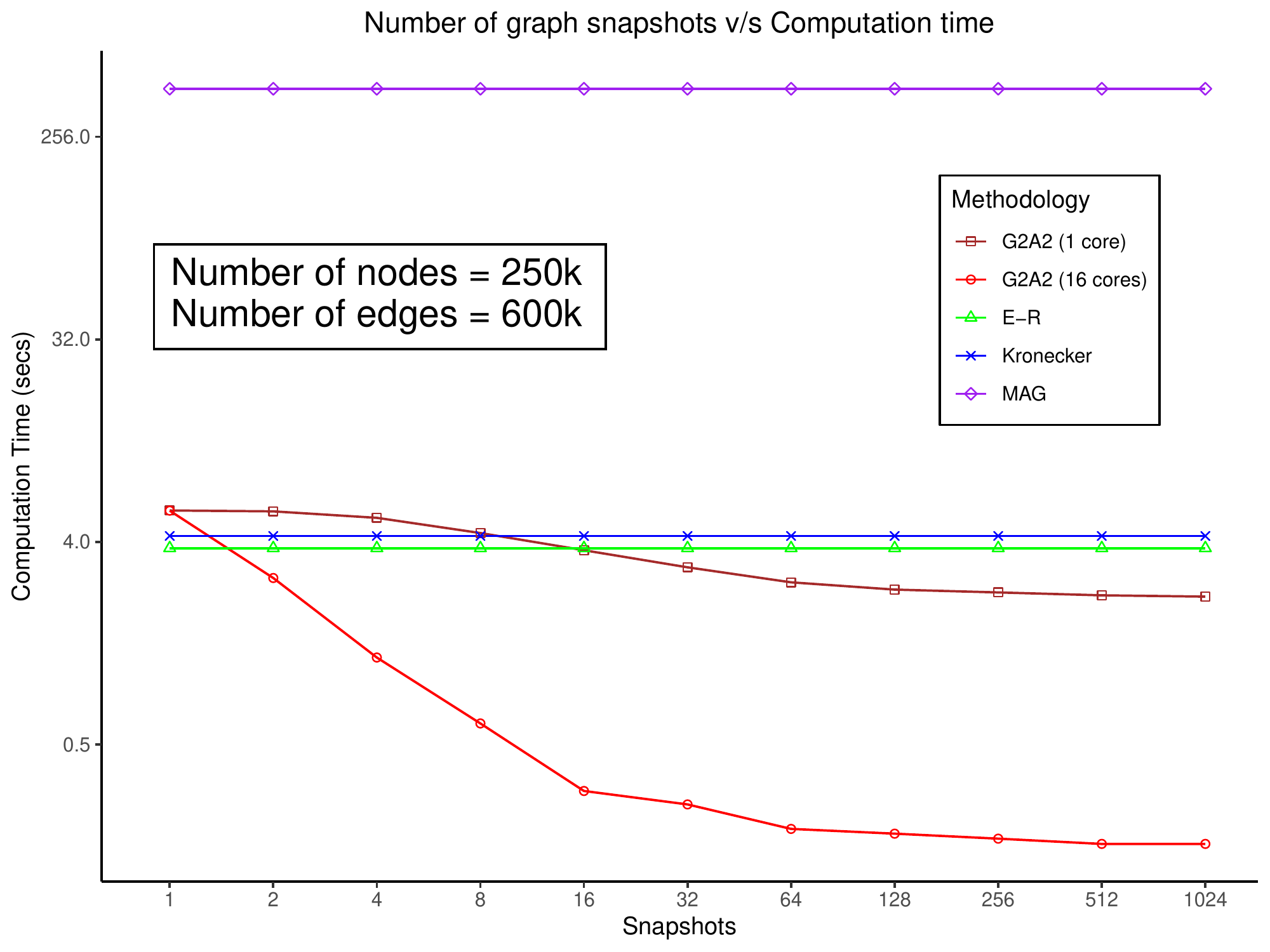}
     \caption{Snapshot analysis}
     \label{fig:SnapshotVsTime}
 \end{subfigure}
    \caption{Computation time of G2A2 vs.\ other baseline models} 
    \vspace*{-10pt}
    \label{fig:Computation Time comparison}
\end{figure*}
\subsubsection{\textbf{Computation time:}}
We compare the computation time of G2A2 against other baseline models across a varying number of edges, number of nodes, and number of snapshots generated, as shown in Figure~\ref{fig:Computation Time comparison}. We only measure the computation time of dynamic bipartite graph generation to keep the comparison fair. The experiments have been performed using Intel(R) Xeon(R) Gold 6130 CPU @ 2.10GHz, with all algorithms running serially. The computation time (in seconds) is on a log scale. G2A2's single-core performance was on par with Kronecker, which is known for its fast computation, even though G2A2 has a higher complexity of graph generation. 
We found that G2A2 computation time is similar to the fastest graph generator algorithm using one core. We see a substantial speedup of G2A2 when running at a higher number of cores.

\subsection{\textbf{Graph library:}}
In addition to a methodology for generating realistic, dynamic, attributed, bipartite graphs and an approach to inject anomalies, this paper provides a graph library with social media graphs, article graphs, and internet traffic graphs generated using the G2A2 methodology. The library will help researchers test their anomaly detection algorithms on graph datasets from multiple domains and with a varying number of nodes, edges, and anomaly percentages. It will also help standardize their anomaly detection algorithms across multiple levels of anomaly percentages.
As per our best knowledge, we are the first to publish a benchmark graph anomaly dataset with various levels of anomaly percentages and varying numbers of nodes and edges. Our graph library can be found here: \href{https://github.com/Sonalj96/G2A2-Graph-Library}{\textbf{https://github.com/Sonalj96/G2A2-Graph-Library}}

\section{Related Work}
In addition to the work mentioned previously in Section~\ref{Introduction}, we discuss a few other approaches relevant to graph generation in this section. For our work, we mainly categorize graph generation methodologies into two: statistical graph generation and deep-learning-based graph generation.
\subsection{Statistical Graph Generation:}
Many of the traditional methods of synthetic graph generation involve random graph modeling. One such model, and the most popular one, is the Erdos-Renyl (E-R) model~\cite{ermodel}. The resultant graphs have minimal utility in terms of real-world applications, which can be traced back to the problem of the E-R model assuming the same probability for every edge in the graph and generating an approximate Poisson distribution. Studies have found that most of the real-world graphs have a small-world effect~\cite{smallworld} and exhibit power-law distribution~\cite{statistic1}, i.e., a skewed distribution. A bulk of such studies focus on static graphs but with increasing interest in time-evolving graphs, works like the forest fire~\cite{forestfire} have presented the characteristics of a time-evolving graph, such as the densification power laws and shrinking diameters. However, there has been a limited effort toward highlighting the distribution of the participating nodes and edges over time. 
\subsection{Deep Learning-based Graph Generation:}
Deep learning-based (DL-based) graph generation has gained significant traction recently. There are a few prominent papers in this area, including~\cite{d2g2,graphrnn}. The advantage of DL methodologies over statistical ones is that the DL-based models learn the intrinsic properties of the graph without explicitly mentioning them. Another advantage is that these models can inherently learn attributes (node or edge). On the other hand, a  disadvantage with these models is that they require a \emph{lot} of data, for which there is a dearth of such publicly available datasets.
Moreover, a majority of DL models only focus on static graphs~\cite{survey2} and fail to generate ground truth from the training graphs.

\section{Conclusion and Future Work}

This paper introduced a new methodology to generate realistic dynamic attributed bipartite graphs with anomalies. Our methodology G2A2 can generate graphs with a realistic degree and time distribution, multiple types of graph anomalies, and attributes similar to an existing non-graph dataset. Our experiments show that G2A2 can easily be parallelized for faster computation and can generate realistic graphs across multiple domains.
In the future, we would like to extend G2A2 to generate graphs with node attributes as well. This will enable researchers to test their anomaly detection algorithms for a wider range of problems. 


    \bibliographystyle{siam}
    \bibliography{ref}

\end{document}